\documentclass[journal]{vgtc}                % final (journal style)
\ifpdf%                                % if we use pdflatex
  \pdfoutput=1\relax                   % create PDFs from pdfLaTeX
  \usepackage{graphicx}                % allow us to embed graphics files
  \DeclareGraphicsExtensions{.pdf,.png,.jpg,.jpeg} % for pdflatex we expect .pdf, .png, or .jpg files
\else%                                 % else we use pure latex
  \usepackage{graphicx}                % allow us to embed graphics files
  \DeclareGraphicsExtensions{.eps}     % for pure latex we expect eps files
\fi%

\graphicspath{{figures/}{pictures/}{images/}{./}} % where to search for the images

\usepackage{microtype}                 % use micro-typography (slightly more compact, better to read)
\PassOptionsToPackage{warn}{textcomp}  % to address font issues with \textrightarrow
\usepackage{textcomp}                  % use better special symbols
\usepackage{mathptmx}                  % use matching math font
\usepackage{times}                     % we use Times as the main font
\usepackage{cite}                      % needed to automatically sort the references
\usepackage{tabu}                      % only used for the table example
\usepackage{booktabs}                  % only used for the table example
\usepackage{amssymb}
\usepackage{caption}
\usepackage{subcaption}
\usepackage{comment}

\onlineid{1440}

\vgtccategory{Research}
\vgtcpapertype{algorithm/technique}

\title{Attribute-based Explanation of Non-Linear Embeddings of High-Dimensional Data}

\author{Jan-Tobias Sohns, Michaela Schmitt, Fabian Jirasek, Hans Hasse, and Heike Leitte}
\authorfooter{
\item
 J.-T. Sohns, M. Schmitt and H. Leitte are with the Visual Information Analysis group at TU Kaiserslautern. E-mail: leitte@cs.uni-kl.de.
\item
 F. Jirasek and H. Hasse are with Laboratory of Engineering Thermodynamics (LTD) at TU Kaiserslautern. E-mail: hans.hasse@mv.uni-kl.de.
}

\shortauthortitle{J.-T. Sohns \MakeLowercase{\textit{et al.}}: Attribute-based Explanation of Non-Linear Embeddings of High-Dimensional Data}
\abstract{Embeddings of high-dimensional data are widely used to explore data, to verify analysis results, and to communicate information. Their explanation, in particular with respect to the input attributes, is often difficult. With linear projects like PCA the axes  can still be annotated meaningfully. With non-linear projections this is no longer possible and alternative strategies such as attribute-based color coding are required. In this paper, we review existing augmentation techniques and discuss their limitations. We present the Non-Linear Embeddings Surveyor (NoLiES) that combines a novel augmentation strategy for projected data (rangesets) with interactive analysis in a small multiples setting. Rangesets use a set-based visualization approach for binned attribute values that enable the user to quickly observe structure and detect outliers. We detail the link between algebraic topology and rangesets and demonstrate the utility of NoLiES in case studies with various challenges (complex attribute value distribution, many attributes, many data points) and a real-world application to understand latent features of matrix completion in thermodynamics.%
} % end of abstract


\CCScatlist{ % not used in journal version
 \CCScat{K.6.1}{Management of Computing and Information Systems}%
{Project and People Management}{Life Cycle};
 \CCScat{K.7.m}{The Computing Profession}{Miscellaneous}{Ethics}
}

\teaser{
  \centering
  \includegraphics[width=\linewidth]{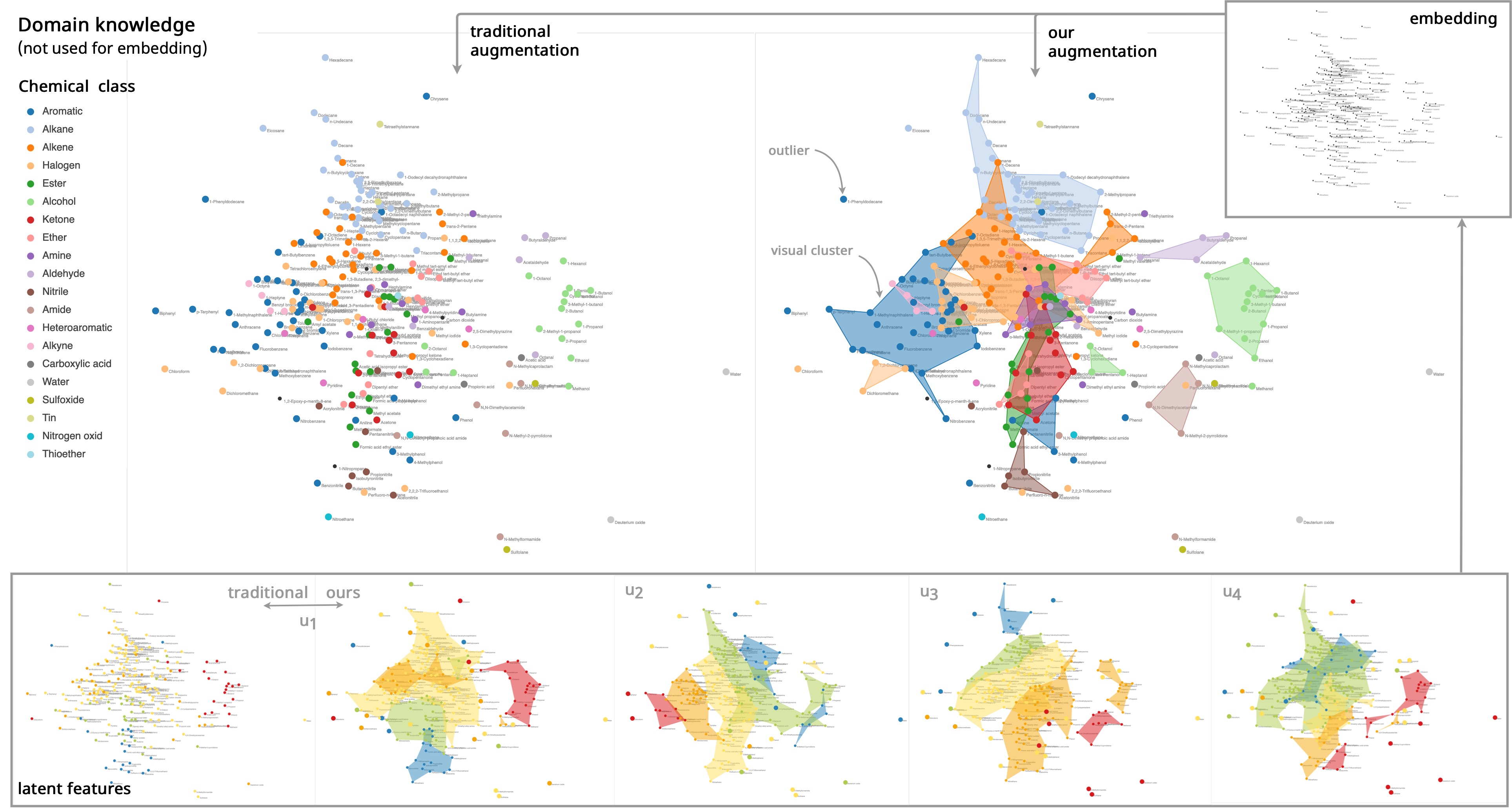}
\caption{Explaining machine learning: The embedding (MDS, top right) shows a non-linear projection of the 4D latent feature space as computed by machine learning for 240 chemical compounds~\cite{doi:10.1021/acs.jpclett.9b03657}. Color-based augmentation helps relate structures in the embedded point cloud to input attributes (bottom, continuous variables) and domain knowledge (top, categorical variables). Our new approach enabled domain scientists detect previously unrecognizable patterns and outliers in their data.%
}
  \label{fig:teaser}
}

\begin{document}


\firstsection{Introduction}

\maketitle

%% \section{Introduction} %for journal use above \firstsection{..} instead
%-------------------------------------------------------------------------
High-dimensional data is omnipresent in the form of tabular data. It occurs in economy, biology, chemistry, political science, astronomy, and physics to only name a few~\cite{7784854, Nonato2019MultidimensionalPF}. For such data a large variety of analysis techniques is offered in modern data analysis libraries ranging from cluster analysis, to regression, to outlier detection, and dimensionality reduction~\cite{scikit-learn}.
The exploratory data analysis principles requests for such techniques to show for each drawn conclusion a raw data plot that convincingly shows such~\cite{tukey1977exploratory}. In many applications this raw data plot is an embedding of the high-dimensional data using methods like linear projections (e.g. PCA) or non-linear techniques like multi-dimensional scaling (MDS) or t-distributed stochastic neighbor embedding (t-SNE). Linear techniques have the advantage that the resulting axes still have meaning, but often they cannot uncover complex structures in high-dimensional space. Non-linear projections often nicely reveal complex structure in high-dimensional space, but no longer offer direct annotation of the projected space. Hence, it is paramount to equip these widely used techniques with mechanisms that help users properly read the projected data and relate original data attributes to the computed features. 
%While the axes still have meaning in linear projections -- though the implications and detailed mapping may be hard to grasp -- non-linear techniques no longer offer direct annotation of space. 
%Hence, it is paramount to understand how to properly read the projected data and relate original data attributes to the computed features. 

\begin{figure*}
  \centering
  \begin{subfigure}{.19\linewidth}
  \includegraphics[height=3.2cm]{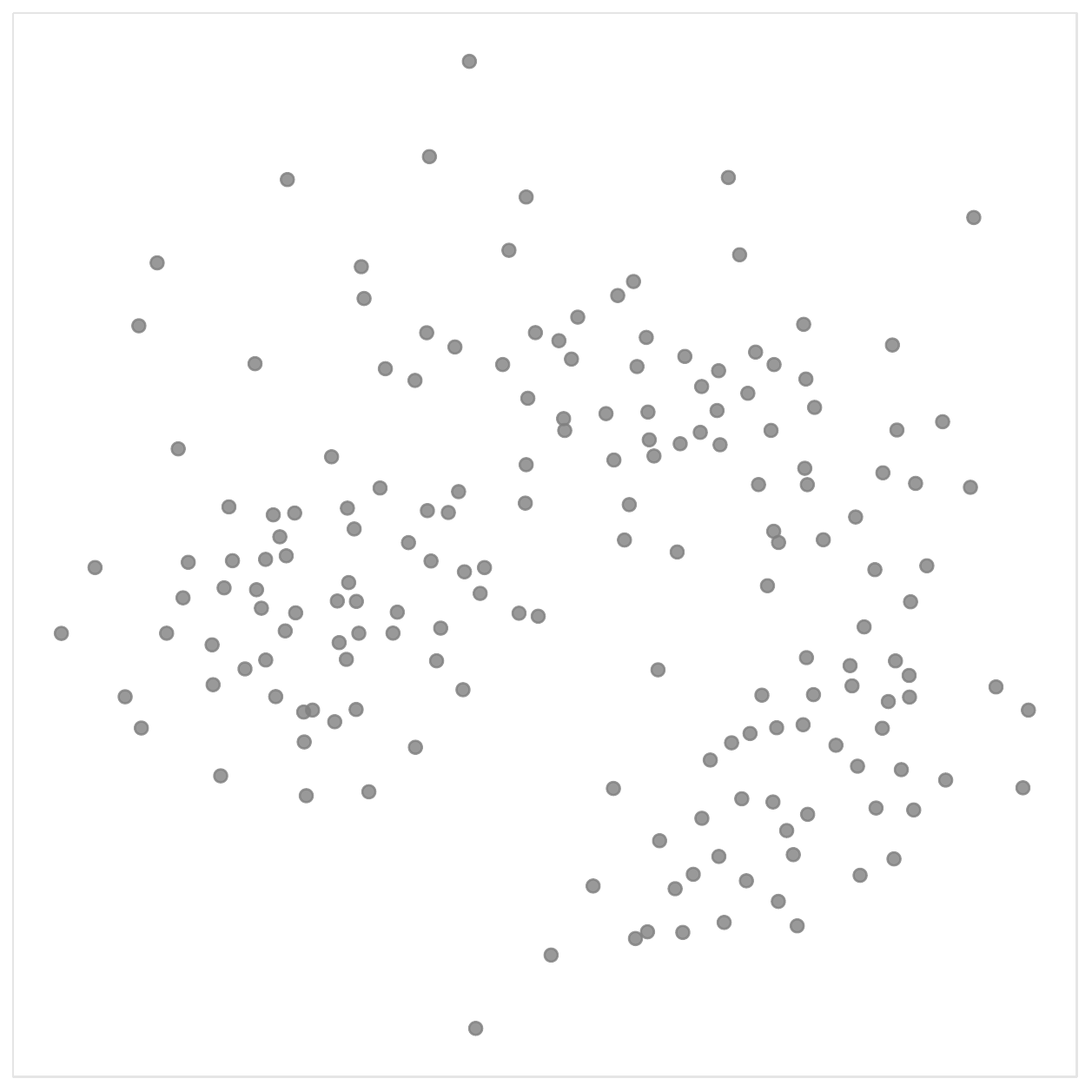}
  \caption{MDS embedding}
  \end{subfigure}
  \begin{subfigure}{.19\linewidth}
  \includegraphics[height=3.2cm]{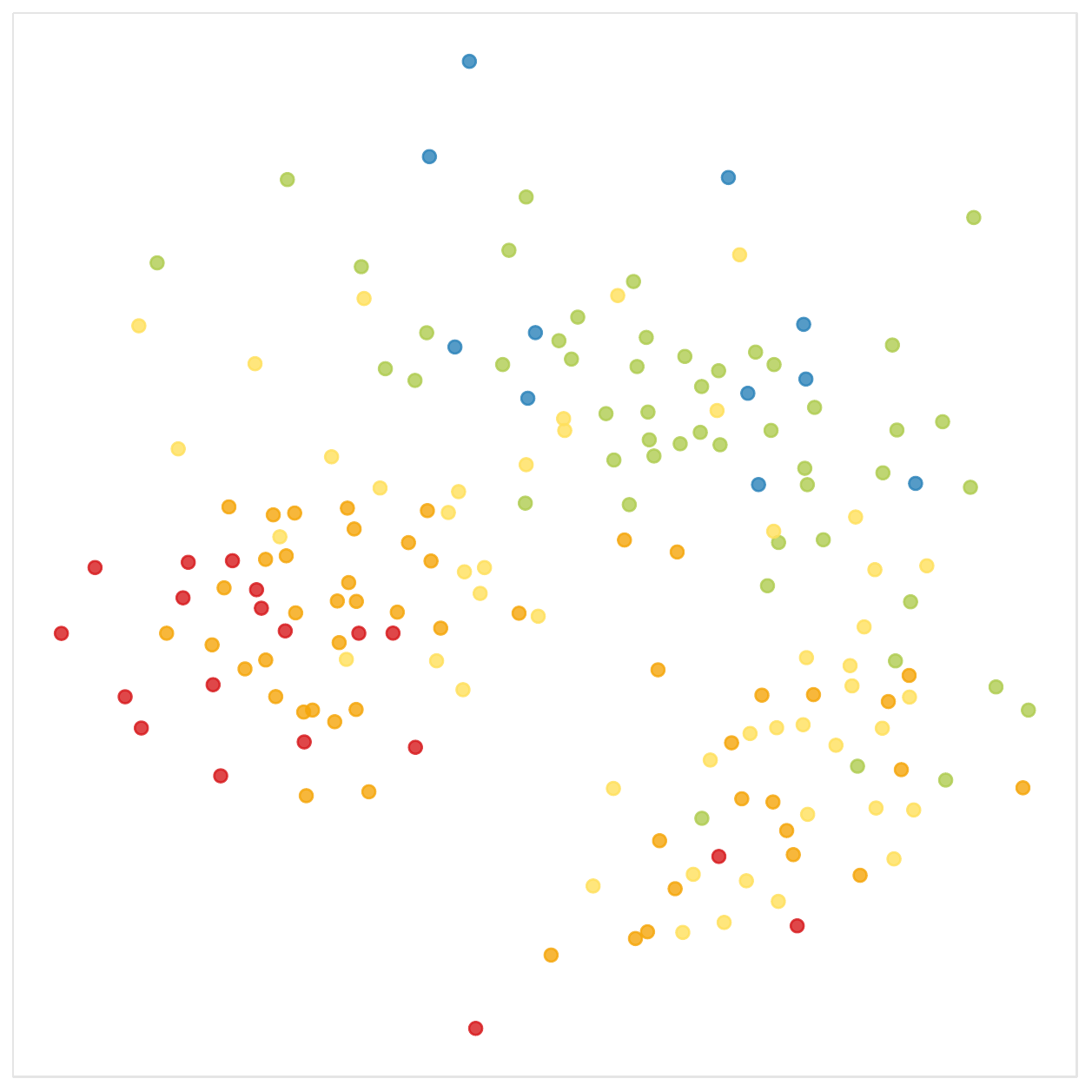}
  
  \caption{Colorcoding}\label{fig:star:glyph}
  \end{subfigure}
  \begin{subfigure}{.19\linewidth}
  \includegraphics[height=3.2cm]{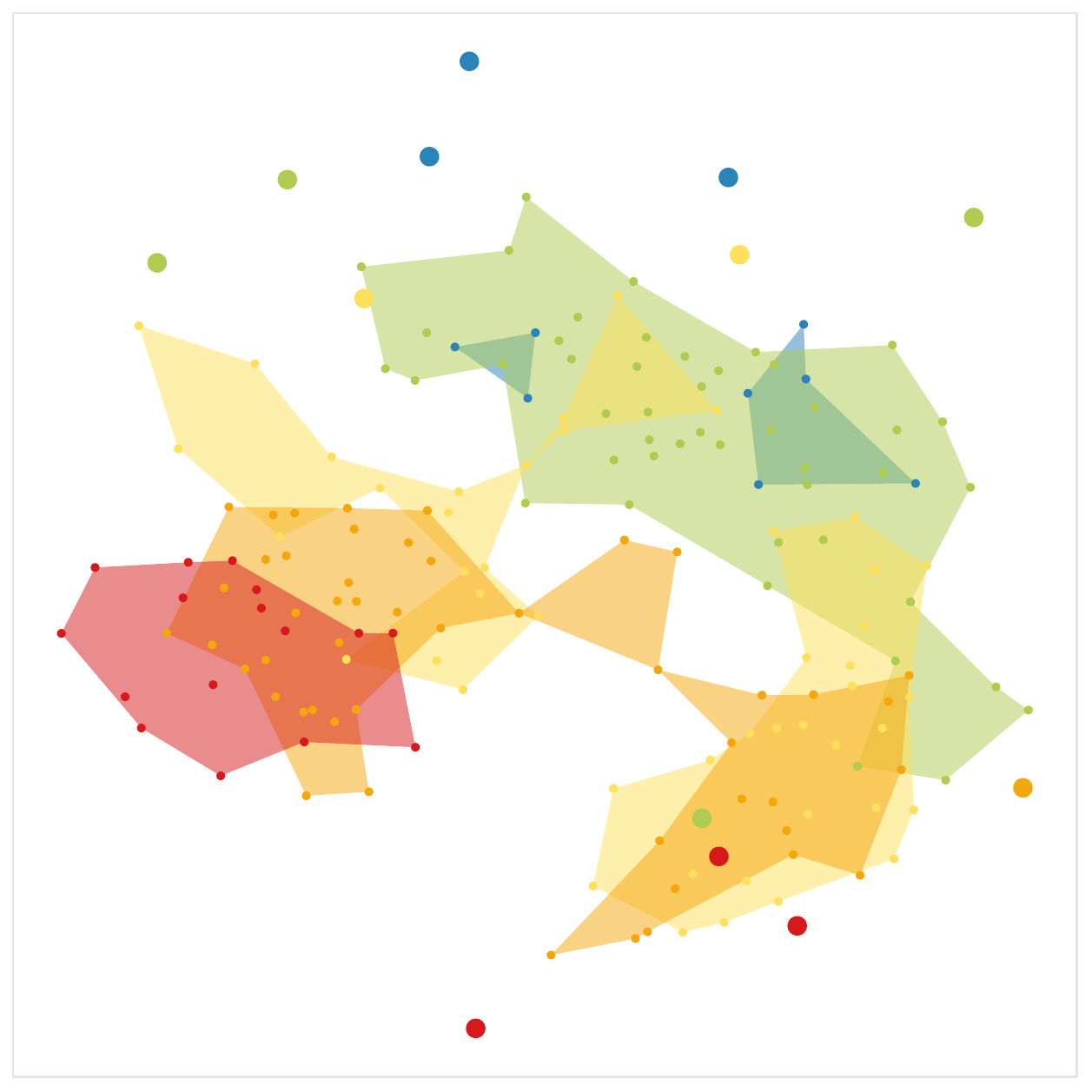}
  
  \caption{Rangesets}\label{fig:star:rangeset}
  \end{subfigure}
  \begin{subfigure}{.19\linewidth}
  \includegraphics[height=3.2cm]{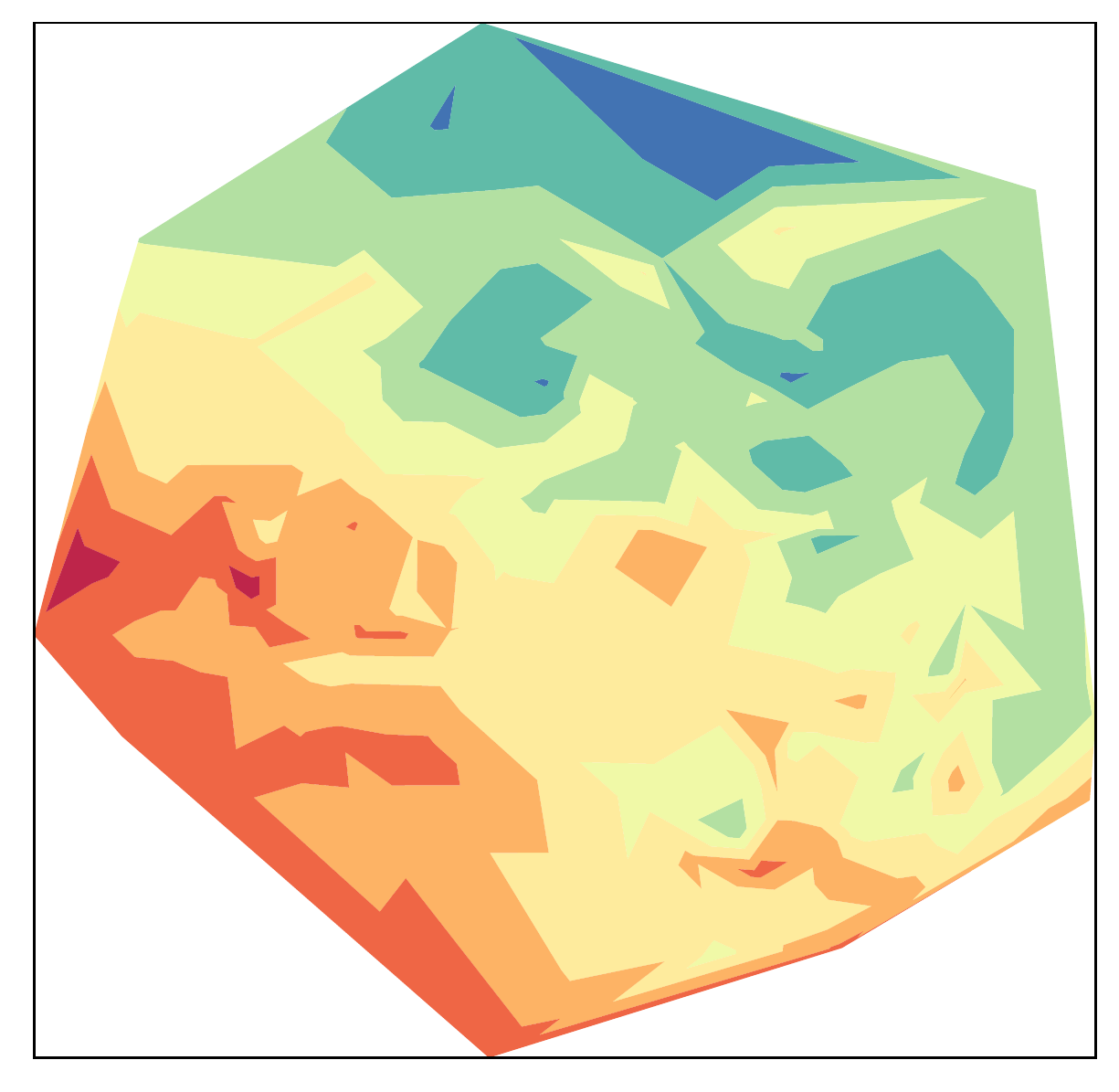}
  
  \caption{Contours}\label{fig:star:contour}
  \end{subfigure}
  \begin{subfigure}{.2\linewidth}
  \includegraphics[height=3.2cm]{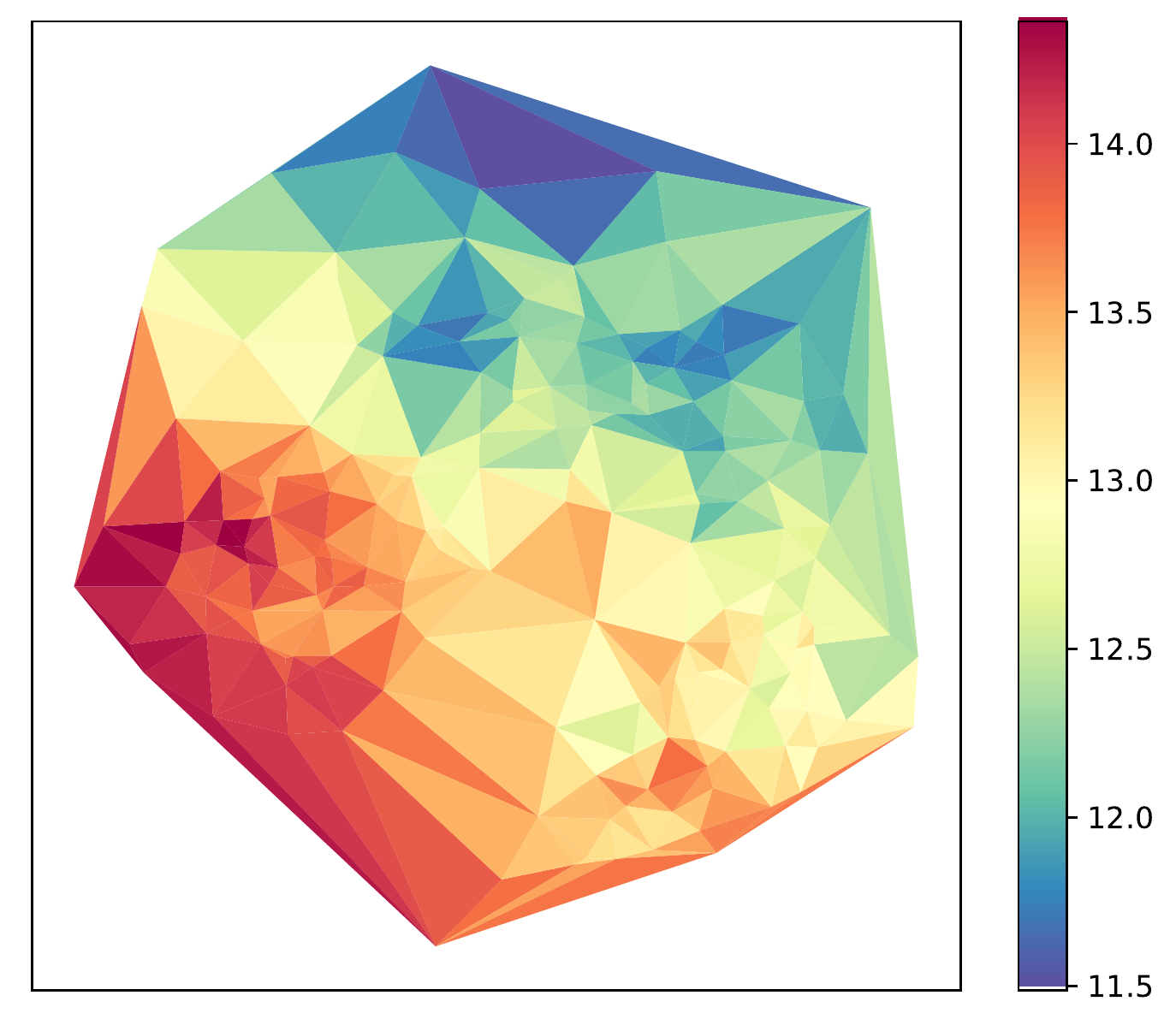}
  
  \caption{Colored triangulation}\label{fig:star:delaunay}
  \end{subfigure}
  \caption{Same data, different chart types: All charts present the same MDS embedding of the wine dataset~\cite{aeberhard1994comparative,Dua:2019} with various augmentation strategies for the attribute alcohol (b-e). Colorcoding (b) correctly represents that data but suffers from occlusion, clutter, and poor outlier highlighting. Field-based techniques (d+e) cannot correctly depict projection ambiguity. Rangesets (c) combine the advantages of both techniques.}
  \label{fig:star}
\end{figure*}

Going through examples in high-dimensional data analysis libraries~\cite{scikit-learn} and recent papers~\cite{7784854}, we observed that the gold standard for this task is still color-coding glyphs and 2D interpolation-based scalarfield reconstruction~\cite{lam1983spatial} as is also confirmed by the survey by Nonato et al.~\cite{Nonato2019MultidimensionalPF}. Fig.~\ref{fig:star} compares these standard techniques for an MDS embedding of the wine dataset~\cite{aeberhard1994comparative,Dua:2019} augmented with the alcohol levels of each sample point. Color-coding assigns each projected data point a color using one of the original attributes (Fig.~\ref{fig:star:glyph}). This technique is easy to implement and to comprehend, but suffers from occlusion and visual clutter and makes outlier detection difficult~\cite{Mayorga2013SplatterplotsOO}.
Scalarfield reconstruction techniques~\cite{lam1983spatial} reconstruct a 2D scalar function for a given attribute, which serves as input to a spatial color-coding (Fig.~\ref{fig:star:contour}~+~\ref{fig:star:delaunay}).
%
%Going through examples in high-dimensional data analysis libraries~\cite{scikit-learn} and recent papers~\cite{7784854}, we observed that the gold standard for this task is still color-coding glyphs and 2d interpolation-based scalarfield reconstruction~\cite{lam1983spatial} (see Fig.~\ref{fig:star} for a comparison) as is also confirmed by the survey by Nonato et al.~\cite{Nonato2019MultidimensionalPF}. Color-coding assigns each projected data point a color using one of the original attributes (Fig.~\ref{fig:star:glyph}). This technique is easy to implement and to comprehend, but suffers from occlusion and visual clutter and makes outlier detection difficult~\cite{Mayorga2013SplatterplotsOO}.
%Scalarfield reconstruction techniques reconstruct a 2D scalar function for a given attribute, which serves as input to a spatial color-coding (Fig.~\ref{fig:star:contour}~+~\ref{fig:star:delaunay}). %Glyph-based color coding easily suffers from occlusion and visual clutter and it is difficult to spot outliers~\cite{Mayorga2013SplatterplotsOO}. 
This results in readily visible spatial patterns, but cannot account 
%Scalarfield reconstruction techniques need to account 
for the fact that data points with different attribute values are projected onto the same position in 2D space. 
%This is solved for example by depicting the local mean, which however, does not correctly reflect the attribute value distribution in 2D space. 
%Nonato et al.~\cite{Nonato2019MultidimensionalPF} survey the state-of-the-art in layout enrichment techniques that augment the embeddings with additional information. Many powerful augmentation techniques were presented in the last decades, focusing particularly on enrichment of the point cloud with additional information such as cluster annotation and automatic labeling. 
%But very little research is present when it comes to correctly representing the underlying raw data. This is a problem that we want to address in this paper. 
%Both strategies have great strengths, but come with major limitations as will be discussed in detail in section~\ref{sec:star}. With our new approach, we want to 
%In our new approach we combine the strengths of each of the two directions
In our new technique, we combine the strengths of these two directions.
Our goals are to design a technique and system that are easy to use and comprehend, that are applicable to all types of embedding techniques, that can be directly integrated into existing analysis pipelines, and that enable the user to quickly and correctly understand attribute value distributions in the embedded data.
We also remark that the goal of this paper is not to explain the projection, i.e., the precise nature of the data transformation from high-dimensional to 2D space, as is done for example in ~\cite{da2015attribute,van2020enhanced} who visually encode least-varying dimensions, but only the final outcome, i.e., the embedding of the data points in the plane.

In this paper, we present NoLiES, an interactive system that enables the user to explore linear and non-linear embeddings with respect to the original attributes. The code is open source\footnote{https://github.com/leitte/NoLiES} and a live demo is provided via MyBinder\footnote{https://bndr.it/2m8ns}. 
NoLiES inherently relies on a novel augmentation technique for multidimensional projections, which we call rangesets. Rangesets, as presented in Fig.~\ref{fig:star:rangeset}, visually group datapoints with similar attribute values into non-convex $\alpha$-hulls. The method allows for outlier filtering and user defined adjustments which is explained in detail in Sect.~\ref{sec:method}.  
%We also present an analysis pipeline that explains how to use the novel rangeset visualization to interpret an embedding .  
To answer questions relating to multiple attributes, an interactive analytics system is required, which will be presented in Sect.~\ref{sec:system}. Several use cases, a real-world application problem of machine learning (ML) in thermodynamics (Sect.~\ref{sec:casestudies}), and feedback from an informal expert user study (Sect.~\ref{sec:discussion}) are provided to demonstrate the capabilities of the novel technique.

In summary, our contributions are as follows:
\begin{itemize}\setlength{\itemsep}{0em}
    \item We review the state-of-the-art of direct augmentation techniques for multivariate projections and highlight strengths and interpretation challenges with these techniques.
    \item We detail a novel visualization technique (rangesets) to augment embeddings. 
    \item We detail the connection of rangesets to algebraic topology and provide first steps on how this theory can be used to further improve augmentations of embeddings.
    \item We present an interactive analysis system and detail the analysis workflow to interpret linear and non-linear embeddings, along with several case studies using examples from ML data bases and real world applications.
\end{itemize}

%Outline:
%\begin{itemize}
%\item Problem: Nonlinear dim red used to represent tabular data. 
%\item Use tabular data to represent relationships between samples and attributes, as well as their relationships among themselves.
%\item Perform Dimension Reduction on data and show embedding.
%\item Explore embedding with interactive tools.
%\item Heike: In an ideal world we would use embeddings of levelsets of the high-dimensional data. With many techniques like MDS this, however, is not possible - we do not have a mapping function, let alone a inverse mapping. We would also need a regression function in high-d space to approximate the sampled function, which causes  problems of its own.\\
%Range-sets are concave hulls of of the intersection of the super- and sublevelsets of the projected data with threshold values according to the range.
%\end{itemize}

\vspace{-4mm}
\section{Related Work}
NoLiES builds upon concepts for augmented multidimensional projections, improved scatterplots, and topological analysis of high-dimensional data, which we will review in the following.
%Non-linear dimensionality reduction techniques are widely used data exploration~\cite{Nonato2019MultidimensionalPF}. The tasks described in this survey are: cluster detection, ... A lot of attention has been paid to the evaluation of the quality of these projections, commonly employing structural error metrics such as density, neighborhood, or .. . 

%Evaluation of dim red for tasks ~\cite{dimara2017conceptual}

\subsection{Augmentation of Multidimensional Projections}
Non-linear dimensionality reduction techniques are widely used for data exploration~\cite{Nonato2019MultidimensionalPF}. Much of the work centers around finding better projections, controlling and communicating error, and automatic detection and visualization of features.
A critical aspect that receives far less support is the interpretation of projected data. Common techniques are (i) direct enrichment, (ii) cluster-based enrichment, and (iii) spatially-structured enrichment \cite{Nonato2019MultidimensionalPF}. 
Direct techniques augment the embedding for example with attribute-based color and text labels to provide additional information~\cite{7784854}. A variety of techniques is presented by Aupetit~\cite{10.1016/j.neucom.2006.11.018} using color-coding on the glyphs~\cite{lehmann2016general, dowling2018sirius, Stahnke2016ProbingPI, https://doi.org/10.1111/j.1467-8659.2012.03108.x}, (hex)bin visualizations for local densities and values~\cite{pagliosa2016understanding}, and multivariate glyph-based approaches. 
%In a similar vain Skupin creates a cartographic representation of the projected data based on a weighted Voronoi tesselation~\cite{974518}. 
All these techniques have in common that they can represent per pixel only attribute values of a single data point. However, most multidimensional projections (linear and non-linear alike) suffer from projection ambiguity, i.e., data points with different attribute values are projected onto the same 2D coordinate, which needs to be communicated. 
%Cluster-based techniques cluster points in the embedding by proximity into non-overlapping regions  and  enrich the visualization with information about the aggregated cluster. 
%Clustering in high-dimensional space and enriching the projection accordingly has also been researched, but has to deal with fragmentation due to the non-linearity of the projection~\cite{https://doi.org/10.1111/j.1467-8659.2012.03108.x}.
The methods closest to our approach are classified as spatially-structured enrichment, which first partition the space using techniques such as Voronoi diagrams or treemaps and afterwards enrich these regions~\cite{https://doi.org/10.1111/cgf.12194, 974518}. Most of these applications, however, aim at a non-overlapping partitioning, which we specifically want to integrate to correctly reflect the nature of the data.

Several techniques aim at the reconstruction of a continuous scalarfield that can be directly visualized using field visualization techniques, e.g., ProbingProjection~\cite{Stahnke2016ProbingPI} or DataContextMap~\cite{Cheng2016ExtendingST}. These spatial techniques directly avoid overplotting and group coherent regions~\cite{Mayorga2013SplatterplotsOO}, but need special strategies to cope with projection ambiguity as will be discussed in more detail in Sect.~\ref{sec:star}. 
%Examples that follow this approach are ProbingProjection~\cite{Stahnke2016ProbingPI} that incorporate heatmaps and the DataContextMap~\cite{Cheng2016ExtendingST} which extends scatterplots to scalar fields.

The third line of research that also follows the concept of a continuous field
%Related techniques that do not operate on a per-datapoint basis try 
tries to recover the non-linear axes or illustrate regions of maximal attribute values. The DataContextMap~\cite{7194836} enriches the embedding with additional data points that locate regions of high attribute values and augment the visualization with additional attribute-based contours on the reconstructed scalar field~\cite{Cheng2016ExtendingST}. 
%Data points close to these reference points feature high values in this attribute, data points farther away feature low values. For data with singular maximal regions in each attribute, this is an excellent and easy to comprehend strategy. For data with multiple maximal regions, this technique can no longer correctly reveal the value distribution. Cheng and Mueller overcome this problem with additional attribute-based contours on the reconstructed scalar field~\cite{Cheng2016ExtendingST}. 
DimReader~\cite{Faust2019DimReaderAL} augments the embedding with non-linear grid lines and prolines~\cite{Cavallo2018AVI} display the non-linear axes. The t-viSNE system~\cite{Chatzimparmpas2020tviSNEIA} presents an analytics tool to explore t-SNE projections~\cite{wattenberg2016how} using for enrichment color-coded glyphs augmented with interactive exploration. An excellent overview of interaction with dimensionality reduction is given by Sacha et al.~\cite{7536217}. %As Sacha et al. state, interactive interfaces are necessary to understand the complex relationship between high- and low-dimensional spaces which cannot be covered in static charts. Our system supports the scenarios feature selection and parameter tuning as presented in the survey.
The techniques in this third category are complimentary to our approach and can be combined with the here proposed rangesets.

%In this paper, we focus on the linking between the projected point cloud and the original variables. Unlike line projection techniques like scatter plots, parallel coordinates or PCA, non-linear techniques no longer have an easy interpretation. This problem has been treated with techniques like the data context map~\cite{7194836}, DimReader~\cite{Faust2019DimReaderAL}, or t-VISNE~\cite{Chatzimparmpas2020tviSNEIA}. A common approach to communicate the structure of the input variable is color coding data points or reconstructing a scalar function (see missing fig). Due to the non-linearity of the projection, this, however, this results in rendering issues as the continuous scalar field "folds over" in the projected version. Christoph mentioned something about Julien and Reeb space here (need to check, edit: "Jacobi Fiber Surfaces for Bivariate Reeb Space Computation" https://ieeexplore.ieee.org/abstract/document/7539583, also a video presentation available: https://vimeo.com/227992719). In this paper, we present techniques that help faithfully communicate scalar value attributes in projected data for large scale data and complex non-linear projections.

\subsection{Improving Scatterplots}
Scatterplot visualization and the treatment of its challenges like overplotting and clutter have been researched in their own right~\cite{Bachthaler2008ContinuousS}. Micallef et al.~\cite{Micallef2017TowardsPO} present techniques to optimize parameter settings for scatterplots like size and opacity to automatically improve the visual results. 
Contours have been applied in a variety of approaches to represent set relationships in scatterplots to improve perception and to lower cognitive load~\cite{5290706, https://doi.org/10.1111/j.1467-8659.2009.01452.x, Cheng2016ExtendingST, Mayorga2013SplatterplotsOO}. Bubble Sets~\cite{5290706} employ density estimates and contour lines to determine outlines for a set. Butterfly Plots~\cite{schreck2008butterfly} use convex hulls and refined convex hulls to enclose data points of the same class. Alpha Shapes, as used in our technique, were previously used by Joia et al.~\cite{joia2015uncovering} to highlight 2D clusters. Simonetto et al.~\cite{https://doi.org/10.1111/j.1467-8659.2009.01452.x} start with a planar graph connecting the points of a set for which a geometric hull is computed. From a theoretical perspective, this approach is similar to ours, though we assume more data points per set, which allows us to go for geometry directly generated through a triangulation. % instead of having to care for many branch-like structures. 
%We use the concept of offset hulls for outlines where we use a state-of-the-art library technique~\cite{shapely}.
%
Mayorga and Gleicher~\cite{Mayorga2013SplatterplotsOO} present Splatterplots that combine density plots with contours to solve the overplotting problem for large numbers of points and at the same time be agnostic to outliers. This approach is conceptually very similar to ours and inspired the presented rangesets. Challenges that we wanted to further improve are locally non-uniform density and cognitive load. A more detailed discussion is given in Sect.~\ref{sec:star}.

\subsection{Topological Methods for High-dimensional Data}
The filtration of the Delaunay triangulation, which is a simplicial complex, allows us to draw from the well researched theory of algebraic topology~\cite{Edelsbrunner1981OnTS, may1992simplicial}. A brief summary of the theory and how it relates to our approach are given in Sect.~\ref{sec:method:topo}. Topological methods have a long history in high-dimensional data analysis~\cite{7784854}. Commonly, they are applied on the high-dimensional data in a preprocessing step to extract relevant structures, e.g. the contour tree~\cite{liu2014distortion} or features~\cite{seifert2010stress, Stahnke2016ProbingPI, rieck2015persistent, rieck2012multivariate}, that help simplify the data and create abstractions that are easier to represent. Topological methods have also been used to control and evaluate the projection process~\cite{rieck2015persistent,10.1016/j.neucom.2006.11.018, doraiswamy2020topomap} integrating Voronoi cells for quality control~\cite{10.1016/j.neucom.2006.11.018}. With this paper, we hope to initiate a connection from the embedding side, so that high-dimensional and two-dimensional topology can mutually augment each other for more insightful data analysis.
%Several papers are concerned with the analysis of the high-dimensional 
%topological approaches for Fiber surfaces and similar - these are extensions of isocontours to higher dimensions, but I think we should be fine with isocontour stuff \cite{Sakurai2017FlexibleFS}

%recovering topology \cite{10.1016/j.neucom.2006.11.018}

%-------------------------------------------------------------------------------------------

\begin{comment}
\section{Requirements Analysis}
Questions we would like to answer: Given an embedding, 
\begin{itemize}
    \item How are attribute values distributed in this point cloud?
    \item Are there attributes that agree with each other, i.e., have a strong correlation (linear or non-linear)?
    \item Why are certain points outliers?
    \item Do clusters in the embedding agree with combinations of attribute values?
    \item Given many attributes. Which ones shall I analyze first?
\end{itemize}

From this we derive the following tasks:
\begin{enumerate}
    \item Represent the distribution of attribute values.
    \item Provide a summary of of the attribute value distribution.
    \item Mark regions and analyze corresponding attribute values.
    \item Adjust visible ranges and the discretization.
    \item Provide guidance to identify relevant attributes for specific tasks.
\end{enumerate}
\end{comment}

%-------------------------------------------------------------------------------------------

\section{Problem Definition and Workflow}\label{sec:workflow}
\begin{figure}
    \centering
    \includegraphics[width=\linewidth]{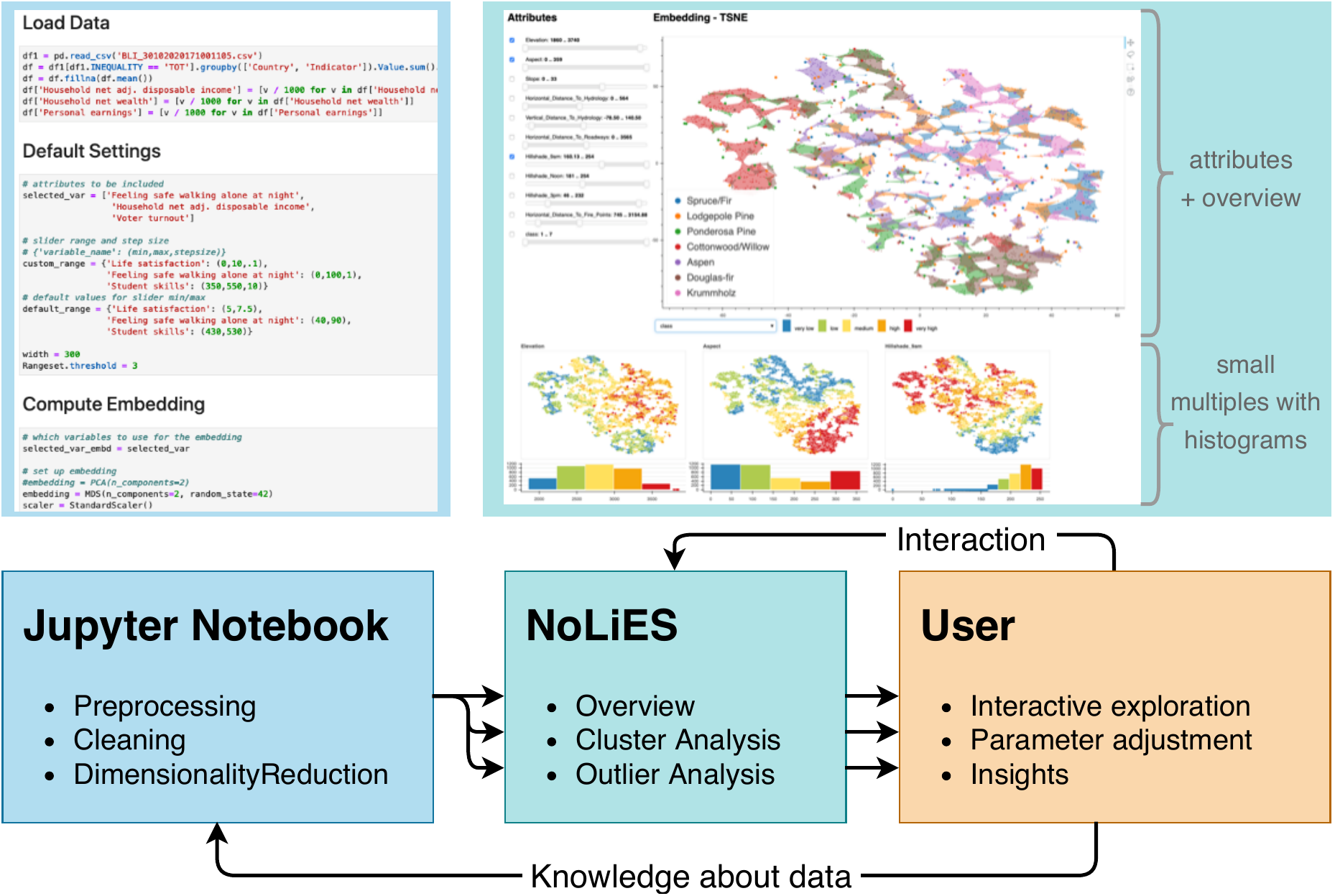}
    \caption{Analysis workflow: NoLiES is implemented in a Jupyter Notebook that is well suited for script-based data preprocessing and can then be served as an interactive web-application. Knowledge that the user obtains during the analysis process, e.g. best parameters or color codes, can be stored in the notebook.}
    \label{fig:workflow}
\end{figure}

NoLiES supports the visual interpretation of dimensionality reduction schemes. Fig.~\ref{fig:workflow} illustrates the three steps of the workflow including screenshots of the two GUI elements. NoLiES is implemented in a Jupyter Notebook~\cite{Kluyver:2016aa} that, when combined with the python panel package~\cite{panel}, comes in two flavors: (Step 1) Scripting and static charts in the Jupyter Notebook and (Step 2) an interactive web-application that supports user interactions. When talking about NoLiES, we primarily refer to the interactive GUI that is used for the data analytics part.

% \begin{figure}
%     \centering
%     \includegraphics[height=3.5cm]{figures/wine_orange_density.pdf}
%     \includegraphics[height=3.5cm]{figures/wine_red_density.pdf}
%     \caption{Density-based contour: For the red and orange points in the wine dataset (high and very high alcohol levels) kde-based contours are computed. Good contour parameters strongly depend on the point density and distribution.}
%     \label{fig:density}
% \end{figure}

%\subsection{Workflow}

\textbf{Step 1 -- Jupyter Notebook} comprises data preprocessing and knowledge storage (Fig.~\ref{fig:workflow}~(left)): In the notebook, the user specifies data loading and cleaning routines. Additionally, they can provide custom parameter values and selections like considered data attributes, slider ranges, filter threshold, or colormaps. In the multi-dimensional projection section, the user can choose their preferred dimensionality reduction method and couple it with appropriate control mechanisms like data scaling and correlation checks.  
%The embedding type is specified and set up including data normalization and similar necessary steps. 
%Default values for the dataset are stored in the notebook, these include optimized parameter ranges, filter parameters for the range sets, and style options like colorcodes.

\begin{figure*}
\centering
\begin{subfigure}[b]{.38\linewidth}
\includegraphics[width=\linewidth]{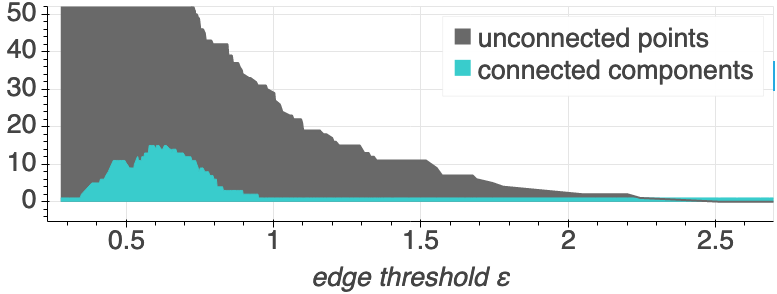}
\caption{Component count wrt to max edge length $\epsilon$}
\end{subfigure}
    %
    %\begin{minipage}{\linewidth}
    \begin{subfigure}[b]{.15\linewidth}
    \includegraphics[height=2.6cm]{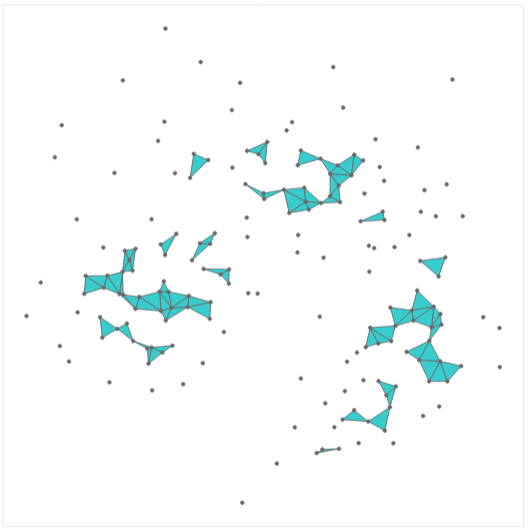}
    \caption{$\epsilon = 0.6$}
    \end{subfigure}
    \begin{subfigure}[b]{.15\linewidth}
    \includegraphics[height=2.6cm]{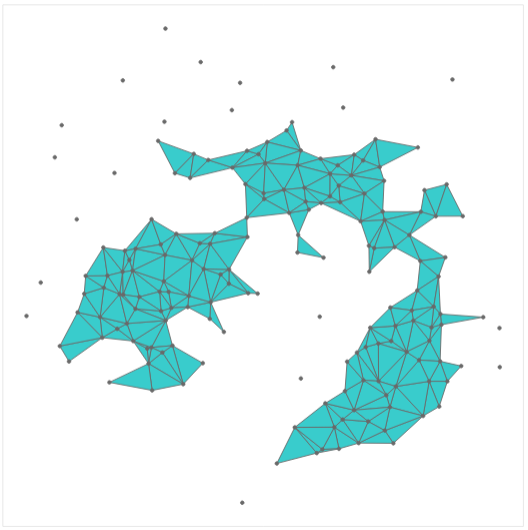}
    \caption{$\epsilon = 1.0$}
    \end{subfigure}
    \begin{subfigure}[b]{.15\linewidth}
    \includegraphics[height=2.6cm]{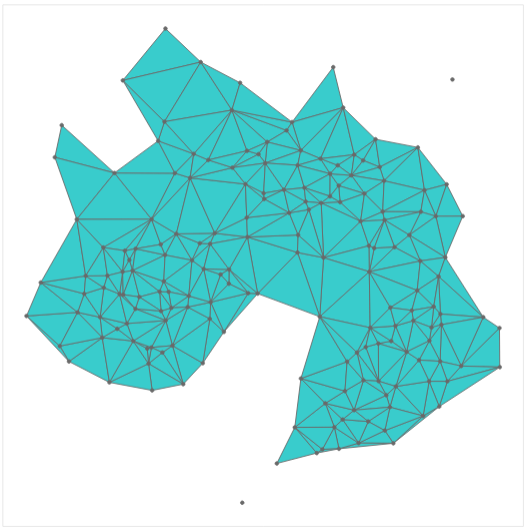}
    \caption{$\epsilon = 2.0$}
    \end{subfigure}
    \begin{subfigure}[b]{.15\linewidth}
    \includegraphics[height=2.6cm]{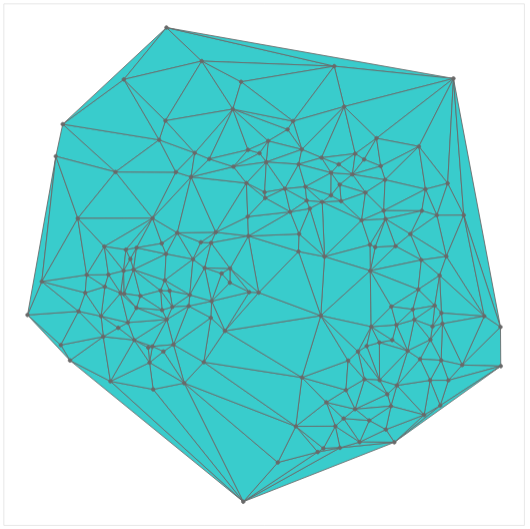}
    \caption{$\epsilon = \epsilon_{max}$}
    \end{subfigure}
    %\end{minipage}
    \caption{Parameter study for the contour parameter: The contour algorithm connects all points that have maximal distance $\epsilon$. (b) Choosing small values results in fragmented contours. (e) Allowing arbitrary distance results in the convex hull.}
    \label{fig:edge_length}
\end{figure*}

\textbf{Step 2 -- NoLiES} for interactive exploration in the GUI (Fig.~\ref{fig:workflow}~(right)): Once the user is happy with the data preprocessing, they change to the interactive GUI view in the browser. In this view, only the chart elements from the notebook are visible and are now linked interactively. The GUI shows the embedded data in a scatterplot and provides overview over included data attributes and attribute value distributions. 

The survey by Nonato and Aupetit~\cite{Nonato2019MultidimensionalPF} gives an excellent overview over analytical tasks performed using multi-dimensional projections. NoLiES supports tasks relating to the two main categories \textit{Explore Dimensions (Axes)} and \textit{Explore Items in Enriched Layout}, i.e., it helps understand the mapped data and explore structures in the projected data respectively.  
%
%Following their taxonomy, we briefly outline how we address three 
%
%Apart from \textit{Map Generation}, they distinguish the following three main task categories: \textit{Explore Dimensions (Axes)}, \textit{Explore Items in Base Layout} and \textit{Explore Items in Enriched Layout}
%
%The tasks supported by NoLiES strongly relate to visual structures and hence,  extend across all major categories as defined by Nonato and Aupetit~\cite{Nonato2019MultidimensionalPF} (\textit{Explore Dimensions (Axes)}, \textit{Explore Items in Base Layout} and \textit{Explore Items in Enriched Layout}). 
In particular, we support the following fine grained tasks (names in italic are tasks as defined by Nonato and Aupetit): (i) \textbf{Explore dimensions} (\textit{Map Synthesized Dimension to Original Dimension, Discover Relation btw. Visual Pattern \& Original Dim.}): We provide a visual link between the position in 2D space and the original data attributes, which explains the attribute value distribution in the projected data. (ii) \textbf{Cluster-based analysis} (\textit{Name Cluster, Discover Clusters in Map, Match Clusters and Classes in Map, Brush in Data Space}): The user observes regions in the embedding that are denser than their surroundings, i.e., clusters. They now want to understand what discriminates the cluster from the surrounding data and what are characteristics that points in the cluster have in common. Small multiples and interactive selections that are shared between all plots support these tasks. (iii) \textbf{Outlier analysis} (\textit{Discover an Outlier in Map, Discover Class-Outlier in Map}): Outliers are points that are unusually far away from other points compared to average point distance. Here, the user would like to understand what sets this point apart from the surrounding points. We support this task by outlier highlighting and the cluster strategies as above.
%
%The three tasks that are commonly pursued in the exploration process are: (a) understanding of the general attribute value distribution, (b) clusters detection and interpretation, and (c) outlier analysis. The general focus of this paper for all tasks is to explain these structure in light of the original variables.
%
%(a) general data distribution: The user observes the distribution of points in the plane as a whole and want to understand of there are designated directions that feature (non-)linear correlations with the original attributes. The goal is to give answers like: Attribute a increases from bottom left to top right.
%
%(b) cluster: The users observes regions in the embedding that are denser than their surroundings, i.e. clusters. They now want to understand what discriminates the cluster from the surrounding data and what are characteristics that points in the cluster have in common.
%
%(c) outliers: Outliers are points that have are unusually far away from other points compared to average point distances. Here the user would like to understand what sets this point apart from the surrounding points.
%

\textbf{Step 3 -- User in the loop} Commonly, the user will need to make adjustments to the default settings during analysis. Hence, the analysis runs in cycles. Additionally, the user can augment the notebook with obtained knowledge to use it in the next run and preserve it for future analysis.

%-------------------------------------------------------------------------------------------
%-------------------------------------------------------------------------

\section{Method}\label{sec:method}
The centerpiece of NoLiES are the proposed rangesets (Fig.~\ref{fig:star}~(center)), which we introduce in this section. We start with a detailed review of existing augmentation strategies, continue with the description of the contour extraction algorithm including references to levelset computation from algebraic topology, and close with the visual encoding.

\subsection{Discussion of Existing Approaches}\label{sec:star}
In the presented rangeset approach, we wanted to overcome  challenges we faced when using state-of-the-art techniques to explore non-linear projections. Starting point was the data displayed in Fig.~\ref{fig:teaser}, which consists of 240 points in 4D. Using standard approaches it was not possible to relate the structure in the point cloud to the input variables or domain knowledge. In this section, we review the three most widely-used concepts for scatterplot augmentation: Glyph-based colorcoding, scalarfield reconstruction, and set-based visualization. For illustration purposes, we use the UCI wine dataset~\cite{aeberhard1994comparative,Dua:2019}, which features the same challenges, but is more widely known and accessible (Fig.~\ref{fig:star}).

%As detailed before,  most of the existing approaches make use of one of two core concepts--glyph-based coloring or scalar field reconstruction and coloring (Fig.~\ref{fig:star}). 
Glyph-based colorcoding (Fig.~\ref{fig:star:glyph}) is easy to implement and gives an intuitive sense of value locations, but suffers from occlusion and overplotting as detailed before~\cite{Mayorga2013SplatterplotsOO}. Assessing the amount of group/color overlap and rapid detection of outliers, however, are difficult.

Fig.~\ref{fig:star:contour}~+~\ref{fig:star:delaunay} represent  scalar fields as reconstructed from the point data. Two popular approaches are nested filled isocontours and triangulation-based~\cite{de1997computational} renderings. 
Both methods are implemented in matplotlib. % The contours are computed using  matplotlib's contour function. Fig.~\ref{fig:star:delaunay} applies colorcoding to the Delaunay triangulation~\cite{de1997computational} of the point set (also implemented in matplotlib). 
While both techniques give a good sense of attribute value distributions, they have difficulties in representing regions that suffer from projection ambiguity, i.e., where points with different attribute values are projected onto each other or close to each other. Here, field-based approaches have to either work with local averages (isocontouring) or create many small color patches (triangulations).

% The rangesets technique in the center (our new approach) combines set visualization~\cite{5290706}  with  outlier highlighting. Comparing the rangesets to the field-based approaches in Fig.~\ref{fig:star}(d+e) we see that (simple) field visualizations cannot truthfully represent data distribution in regions where data points with varying attribute values overlap -- look, for example, at the intersection of the orange and red contour. Fig.~\ref{fig:star:contour} uses the matplotlib contour function which renders levelsets in a reconstructed scalarfield. Fig.~\ref{fig:star:delaunay} applies colormapping to a Delaunay triangulation~\cite{de1997computational} of the point set which results in rapidly changing color patches in the overlapping regions. %This technique gives locally correct results but sacrifices the improved perception as provided through set visualizations as regions with varying attribute values result in many small triangles with varying colors next to each other -- overlap is per definition not intended.
% Nevertheless, region-based augmentation is a highly appealing approach as it visually groups the data and eases chart reading~\cite{Mayorga2013SplatterplotsOO}.

A third concept that is used particularly for categorical attributes on scatterplots are set-based visualizations~\cite{Nonato2019MultidimensionalPF}. The two primary directions in which the outlines of sets can be obtained are geometric/algebraic approaches and statistical approaches.
The geometric approaches operate on a simplicial complex (graph or triangulation) and derives the boundary from this construct through filtering or additional geometric operations like dilation. The convex hull of a set of points is an example of a geometrically created boundary.
The statistical approach relies on a density estimate for which an isocontour is drawn. Both approaches, statistical and geometric, rely on parameters that control the outlier filtering. In the statistical case, this is achieved through the isovalue of the boundary contour. A known problem is the challenge in handling point clouds with locally varying density, which are treated, for example, with adaptive-KDE-methods~\cite{van2003adaptive}, which, however, are hardly implemented in data analysis software packages. 
%Figure~\ref{fig:density} illustrates multiple isocontours for a subset of points -- our goal is to render multiple contours for datapoints in given value ranges. In the left image the kernel-density-estimate (kde) was computed for the orange points and four contours for different isovalues are rendered. The right image shows the same procedure for a different selection. These images illustrate the challenges that one commonly faces when working with density estimates. Embeddings usually have varying point density across the plot, i.e., there are clusters where points are densely packed and others with a more lose clustering (Fig.~\ref{fig:density}(right)) which cannot be accounted for easily. The problem is treated with adaptive-kde-methods~\cite{van2003adaptive} which are currently not widely implemented in software libraries. A more important challenge is the difficult selection of an appropriate isovalue which again would be different for each set of points (compare left and right figure).
Geometric approaches, on the other hand, are commonly controlled by geometric criteria like maximal distance to identify outliers.%of two points belonging to one set as used in rangesets. The quality of the extracted structures (contours and outliers) strongly depends on the quality of the data

In conclusion, we want to state that all methods have use cases where they are best suited. Glyph coloring is easy to implement and works very well with a limited number of data points. Statistical approaches carry the interesting notion of probability that points may be located in a certain area of the plot and scalar-field techniques can handle an arbitrary number of color-levels if this is requested. For the rangesets presented in this paper, we chose the geometric set-based approach as it uses the strengths of spatial augmentations, can handle projection ambiguity, and is intuitive to control by a single distance parameter with good default value heuristics.%

\subsection{Computation of Contours}
\begin{figure}
    \centering
    \includegraphics[width=\linewidth]{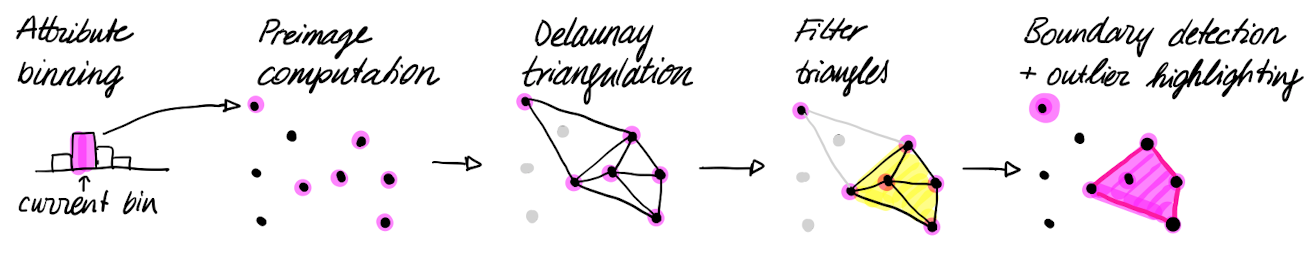}
    \caption{Contour computation: The algorithm consists of five steps and the key tasks are highlighted in the illustrations.}
    \label{fig:algo}
\end{figure}

\begin{figure}
    \centering
    \includegraphics[width=.45\linewidth]{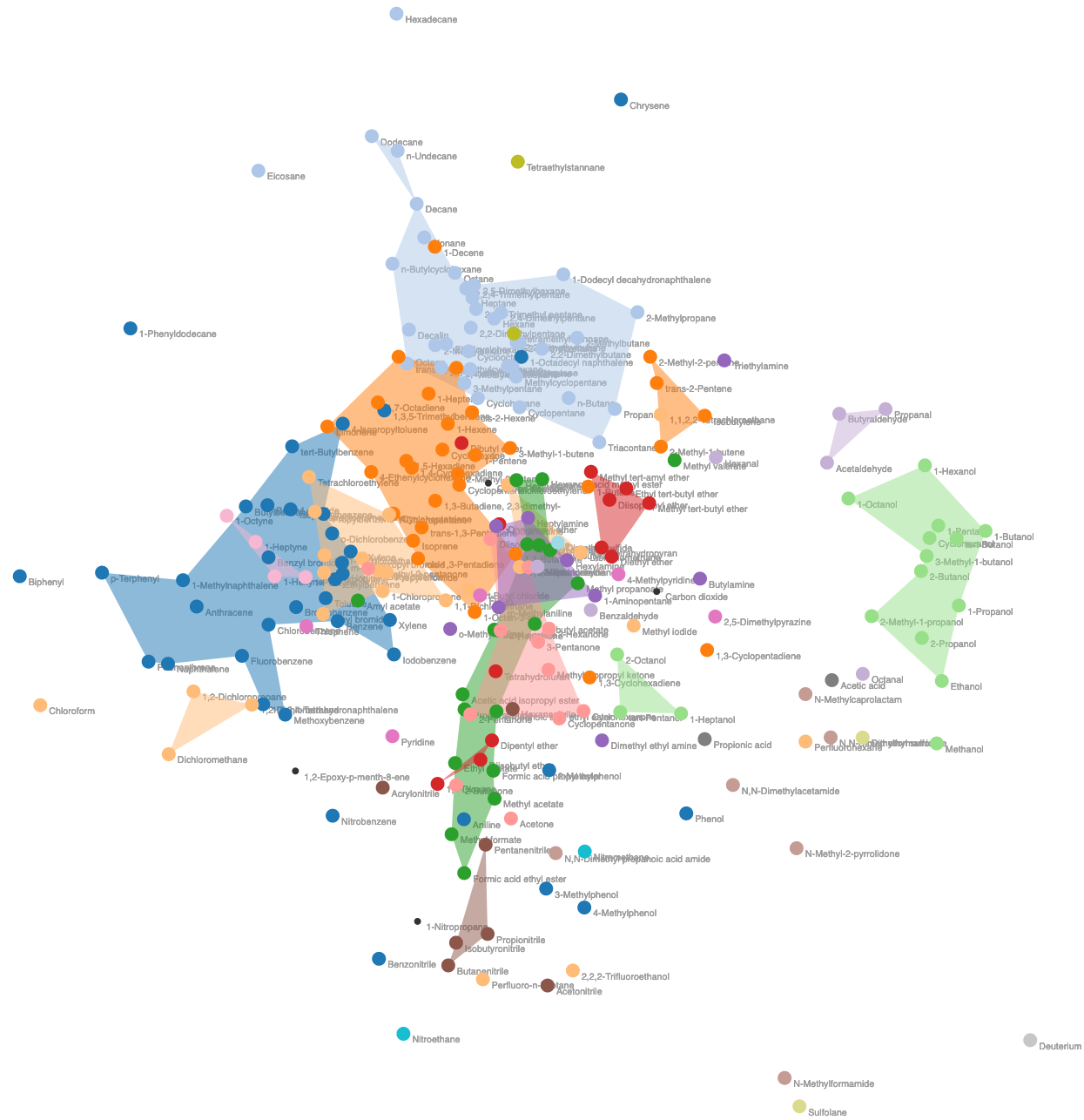}\hspace{4mm}
    \includegraphics[width=.45\linewidth]{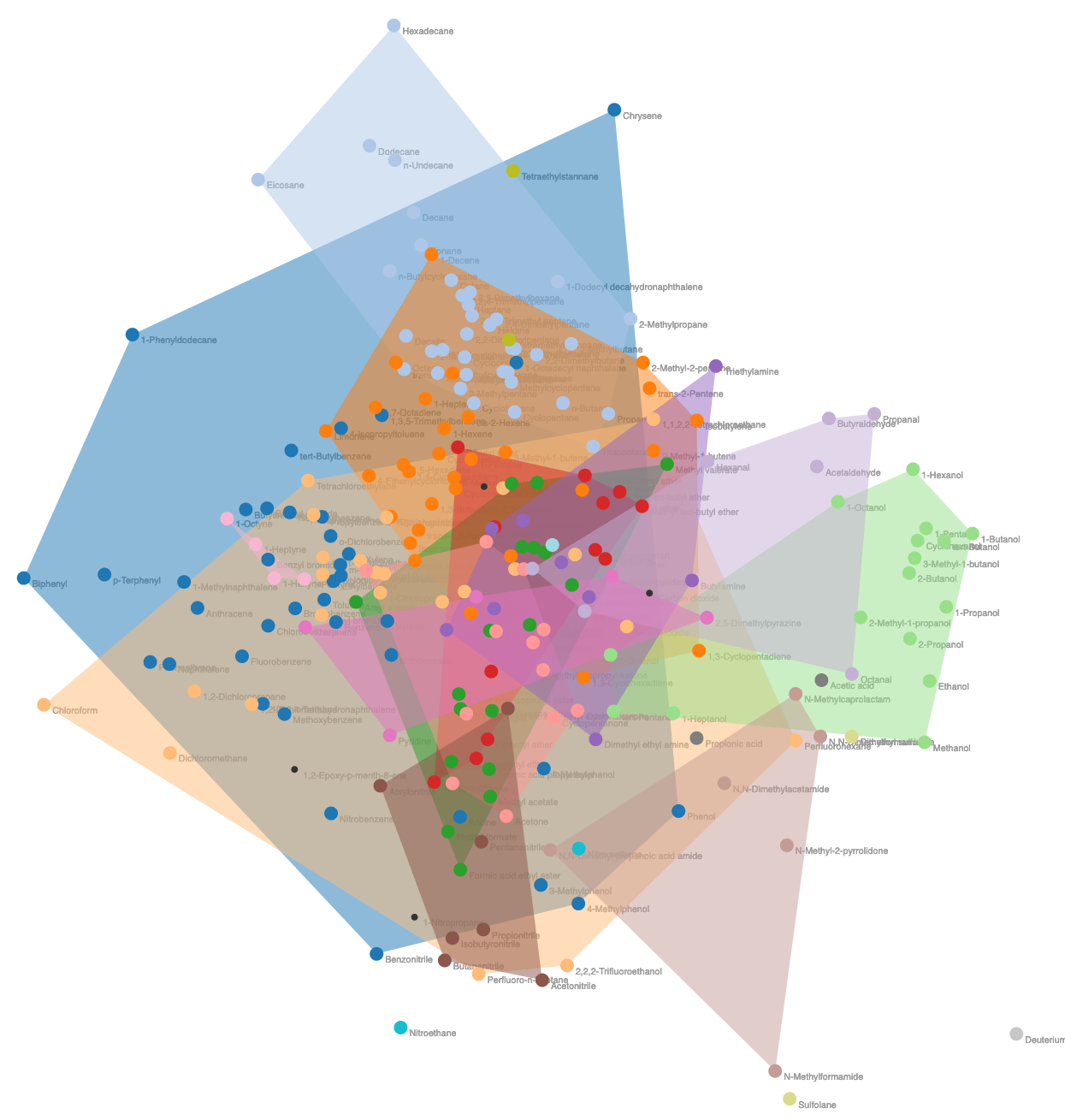}
    \caption{Effects of the distance parameter $\epsilon$ on final layout: The thermodynamics embedding is augmented with color for 20 chemical classes. Small values for $\epsilon$ (left: 0.75) result in core regions of the classes. Large values (right: 10) result in convex hulls that strongly overlap.}
    \label{fig:thermo_epsilon}
\end{figure}

Rangesets use geometric contours to visually group data points with similar values. For categorical attributes  a contour is drawn per category (see Fig.~\ref{fig:thermo_epsilon}). For continuous variables, we first discretize the value range of the attribute and then draw a contour per bin, i.e., for a range of attribute values (see Fig.~\ref{fig:discretization}). In the following, we detail the algorithms used to compute the contour(s) per value bin and to filter outliers. 
%
%
%The process of contour computation for a given attribute is illustrated in Figure~\ref{fig:algo}. We assume that a discretization of the value range of an attribute is given, i.e., we are given a histogram with bin ranges and bin counts. The algorithm consists of the following steps:
The final algorithm then consists of the following five steps as illustrated in Fig.~\ref{fig:algo}:
(i) select an attribute bin, (ii) filter data points in the respective range, (iii) compute Delaunay triangulation of the filtered points, (iv) remove triangles with unwanted properties, (v) compute the boundary of the triangulation and find all points in the current range that do not belong to the contour for highlighting.

%We assume that a discretization of the value range of an attribute is given (step i), i.e., we are given a histogram with bin ranges and bin counts. The choice of discretization will be discussed in more detail in section~\ref{sec:method:discretization}. For each bin we want to illustrate the preimage in the scatterplot, i.e., highlight all data points that contributed to a given bin, following the goals (sec.~\ref{sec:intro}) and design criteria (sec~\ref{sec:star}) as discussed before. We detail the algorithm for a single bin and discuss the combination of multiple bins in section~\ref{sec:method:encoding}.

%The steps of the rangeset algorithm are as follows (Figure~\ref{fig:algo}): (i) select an attribute bin, (ii) filter data points in the respective range, (iii) compute Delaunay triangulation of the filtered points, (iv) remove triangles with unwanted properties, (v) compute the boundary of the triangulation and find all points in the current range that do not belong to the contour for highlighting.

%-------------------------------------------------------------------------

\subsubsection{Non-Convex Hulls}
%In the following, we describe the algorithm to compute the preimage for a single bin. Combinations for multiple bins are discussed in section~\ref{sec:method:encoding}. 
Given a set $S$ of $n$ points in the plane ($n$ being a positive integer), we are looking for a hull that tightly encloses the points of $S$ and allows for filtering of outliers. 
Non-convex or minimum area hulls~\cite{arkin1998minimum} feature these properties. Following the discussion for the choice of a non-convex hull for scagnostics~\cite{wilkinson2005graph}, we use $\alpha$-hulls as they can be computed efficiently and allow for filtering of outliers.
The $\alpha$-hull of $S$ is the intersection of all closed complements of discs with radius $-1/\alpha$ that contain all the points of $S$ for arbitrary negative reals $\alpha$.
As shown in~\cite{Edelsbrunner1981OnTS}, $\alpha$-hulls can be computed efficiently from the Delaunay triangulation~\cite{de1997computational} of the point set $S$ by excluding triangles that contain an edge whose length exceeds a given threshold $\epsilon$. For the distance computation we use the Euclidean distance $d(\cdot,\cdot)$. 
The filtered Delaunay graph $\mathcal{D_\epsilon}$ now has the vertex set $V=\{0,1,...,n-1\}$ and an edge set of $E=\{(u,v) \in V\times V | d(u,v)\leq \epsilon\}$, i.e., two vertices $u$ and $v$ are connected if and only if their distance is less than or equal to the selected distance threshold $\epsilon$.
The parameter $\epsilon$ controls the ``tightness'' of the computed contour. Large values include all edges of the Delaunay triangulation resulting in the convex hull~\cite{de1997computational} of the point set. Very small values will result in a strong fragmentation. 

An example that illustrates the progression of the contours with increasing $\epsilon$-values is shown in Fig.~\ref{fig:edge_length}. Here, all points of the wine dataset are included in the contouring process and the resulting contours are shown for increasing $\epsilon$-thresholds. Note how the set of contours changes from small fragments, over a smooth tight boundary, to the full convex hull. %Similar observations were made for all datasets and follow from the 
The choice of the "tightness" of the boundary has strong effects on the final augmentation for the embedding where multiple contours are drawn. An additional complex example is given in Fig.~\ref{fig:thermo_epsilon} for the thermodynamics dataset where 20 categorical classes are used to explain the embedding. For small $\epsilon$-values, the contours for each class focus on dense core regions (left, $\epsilon = 1$). For large values, the contours overlap strongly and are no longer helpful (right). Finding a good $\epsilon$-value is an important aspect of the algorithm and will be discussed in the next section.

%Using the convex hull for each bin results in excessively large contours that overlap in many regions (see Fig.~\ref{fig:thermo_epsilon} for two examples). Hence, we need to find an appropriate edge threshold $\epsilon$ which will be discussed in the next section.

%-------------------------------------------------------------------------

\subsubsection{Topological Filtration to Control $\epsilon$}\label{sec:method:topo}
To better understand the progression of the size of the contour, we now look into algebraic topology. The above outlined algorithm induces a filtration of the simplicial complex, i.e., the subcomplexes $\mathcal{D}_{\epsilon_1}$ and $\mathcal{D}_{\epsilon_2}$ form a nesting hierarchy with $\mathcal{D}_{\epsilon_1}$ being a subset of $\mathcal{D}_{\epsilon_2}$ if and only if $\epsilon_1 \leq \epsilon_2$. On the operational level, this results in the nice property that increasing the threshold $\epsilon$ can only add triangles to the non-convex hull, but never remove them. For $\epsilon = 0$ no contour is created and each data point is an outlier and for $\epsilon = \epsilon_{max}$, where $\epsilon_{max}$ is the longest edge in the Delaunay graph, we obtain the convex hull of the set of points in the plane. For $\epsilon$-values within this range, the number of outliers will decrease and the number of contours will vary. Algebraic topology is a research field that studies topological properties and changes of the simplicial complex under filtrations. For our analysis, we currently consider the zero Betti-number $b_0$, which captures the number of connected components. Another helpful characteristic is the first Betti-number, which counts the number of holes in polygons.

To obtain a better understanding of the effects of $\epsilon$, we additionally provide a topological summary. In simply connected domains like ours, this information is often presented as a contour tree~\cite{carr2003computing}, which depicts the merging of the contours as $\epsilon$ changes. As the more critical information for our application is the number of outliers and connected components for any $\epsilon$, we provide this information in an area chart. You can think of this chart as a marginal of the merge tree differentiating between connected components containing exactly one and those containing multiple vertices.

An example of the topology chart is given in Fig.~\ref{fig:edge_length}. Connected components with more than one vertex are depicted in blue and outliers in gray. The plot is truncated at the top as in the beginning $\epsilon \in (0,0.7)$ most data points are outliers. Alternatively, the user can switch to a logarithmic y-axis. The graph presented here is typical for all datasets that we investigated. In the beginning, with $\epsilon = 0$,  all datapoints are outliers. With increasing $\epsilon$-values many small contours form ($\epsilon \in (0.2,0.7)$), which eventually merge to larger more stable contours. Starting from $\epsilon = 1$, we only have one contour and the threshold parameter only controls the number of outliers. Note that the presented $\epsilon$-values are not universal as they refer to edge length as computed by the embedding. Scaling the embedding can help, but one still has to account for varying aspect ratios and point densities due to number of points.

Wilkinson et al.~\cite{wilkinson2005graph} propose a default value for $\epsilon$ based on edge lengths in the minimal spanning tree (MST): 
$$\epsilon = q_{75} + 1.5\cdot(q_{75} - q_{25})$$ where $q_{75}$ is the 75th percentile of the MST edge lengths and the expression in the parentheses is the interquartile range of the edge lengths. It is important to note that the minimum  spanning tree of a set P of point sites (in any dimension) is a subgraph of the Delaunay triangulation. Hence, we use this criterion to provide a default value for the filter threshold $\epsilon$. In most practical applications we found this value too restrictive and manually raised the threshold to obtain smoother contours. An example is given in Fig.~\ref{fig:edge_length} where the suggested $\epsilon$ is 0.96 (similar to Fig.~\ref{fig:edge_length}(c)), but was overwritten with $\epsilon=2$ for the analysis.

%-------------------------------------------------------------------------

\subsubsection{Discussion of Triangle Filter Criteria}
The above described algorithm demonstrates how topological analysis can be used to control the contours and we used maximal edge length for the distance function. Alternative choices, however, are possible. The three widely used local criteria when controlling triangulation quality are edge length, triangle area, and inner angles. Optimizing inner angles is already ensured through the Delaunay triangulation, which creates an angle optimal triangulation, i.e., from all possible triangulations the one with largest smallest angle is chosen~\cite{de1997computational}. This ensures that long spiky triangles are avoided as much as possible.

\begin{figure}
    %\centering
    %\includegraphics[width=\linewidth]{figures/wine_filtration.png}
    \includegraphics[width=.45\linewidth]{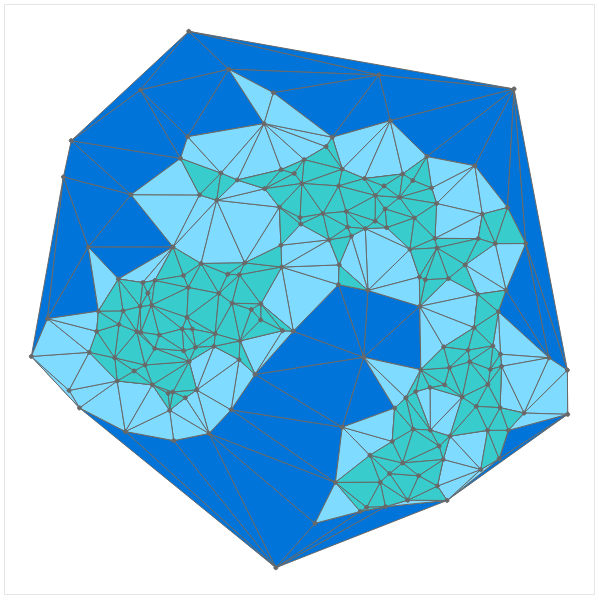}
    \includegraphics[width=.45\linewidth]{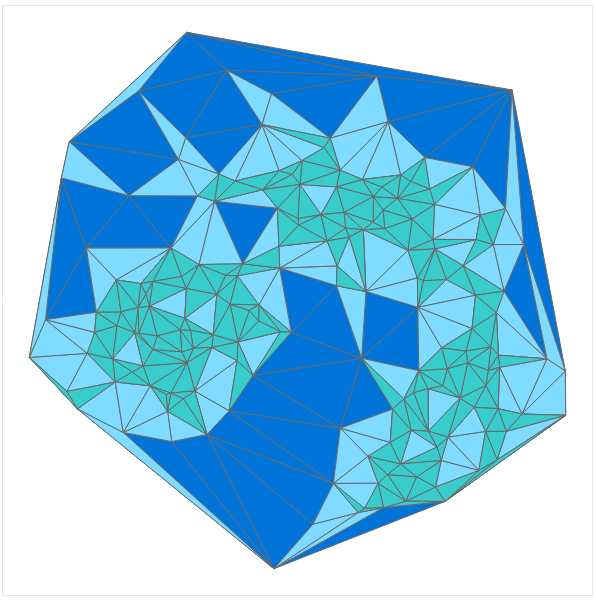}
    
  %\begin{minipage}{0.48\linewidth}
    %\includegraphics[width=.45\linewidth]{figures/wine_filtration_areaplot.png}
    %\medskip
    %\hfill
    %\includegraphics[width=.45\linewidth]{figures/wine_filtration_edgeplot.png}
  %\end{minipage}\hfill
  %\begin{minipage}{0.5\linewidth}
  %\end{minipage}
  
    \caption{Comparison of filtration attributes: (left) Filtration based on edge length with thresholds of 0.9, 1.55, $\epsilon_{max}$. (right) Area-based filtration with thresholds 0.2, 0.5, $\Delta_{max}$, which features unwanted properties like spiky triangles and holes.}
    \label{fig:filtration}
\end{figure}

In the above algorithm, we used edge length for the filtration. An alternative choice of distance metric for the filtration that we explored during the development of NoLiES is the area of triangles, i.e., triangles and their bounding edges are included in $\mathcal{D}_\epsilon$ if they do not exceed the threshold size $\epsilon$. The goal was to exclude large triangles that cover space without additional points inside the triangle. A comparison of the two distance metrics can be seen in Fig.~\ref{fig:filtration}. This choice of metric, however, proved to be much harder to control and resulted in unexpected holes in the hull and long spiky triangles not being removed. %Hence, we use the edge-length-based filtering as default criterion.

%-------------------------------------------------------------------------

\begin{table}
\begin{center}
\begin{tabular}{ |c|c|c|c|c|c|c|c| } 
 \hline
 n points & 1k & 2.5k & 5k & 10k & 15k & 20k & 50k \\ 
 \hline
 UMAP & 4.6 & 5.0 & 3.8 & 6.1 & 4.9 & 6.2 & 14.4\\ 
 MST & 0.2 & 1.3 & 6.0 & 28.7 & 75.9 & 178.4 & ---\\ 
 Rangeset & 0.6 & 3.3 & 7.3 & 11.3 & 20.5 & 35.5 & 57.7\\ 
 \hline
\end{tabular}
\end{center}

    \caption{Timings on the covertype dataset: A subset of the 500k points was sampled and the three important algorithmic steps (embedding: UMAP, error computation and $\epsilon$-estimation: minimal spanning tree MST, rangeset computation) were timed. Timings are given in seconds.}
    \label{fig:table:timings}
\end{table}

\subsubsection{Scalability and Stability}
\begin{figure}
    \centering
    \includegraphics[width=\linewidth]{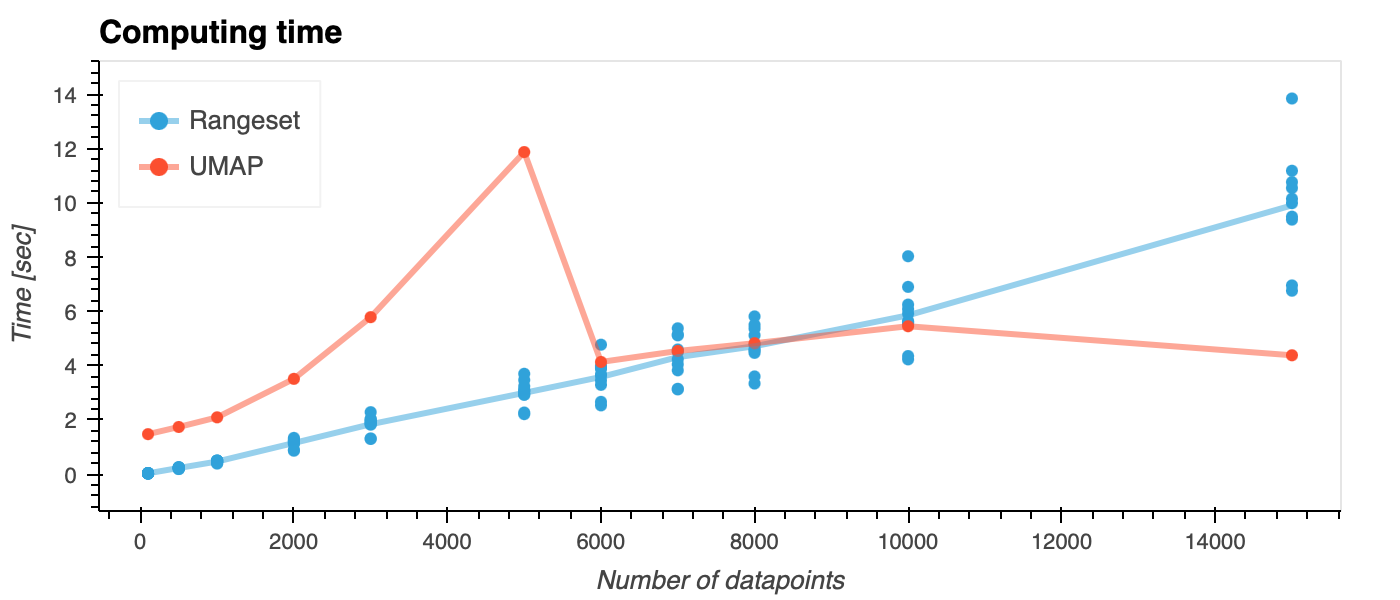}
    \caption{Benchmark for the covertype dataset: The UMAP embedding is a stochastic approach designed for handling large-scale datasets. Rangesets scale linearly with the number of datapoints. Timings are given for the rangeset of one variable per bin, i.e., the total time for one chart is the sum of times as given in table~\ref{fig:table:timings}.}
    \label{fig:benchmark}
\end{figure}

Algorithms from algebraic topology have great analytical power, but are known to be computationally expensive~\cite{bauer2021ripser}. In our practical examples, we found the system to be interactive for datasets with up to 1,000 data points. Up to 5,000 data points required acceptable waiting times of a few seconds. We conducted a systematic study to assess running times for larger datasets. For the analysis we sampled the covertype dataset~\cite{Blackard1999ComparativeAO} which consists of 500k data points. For the embedding we used the UMAP algorithm~\cite{mcinnes2018umap} which was able to project the data within a couple of second for various sample sizes (Table~\ref{fig:table:timings}). Assessing projection quality requires the computation of the minimal spanning tree (MST - second row)~\cite{motta2015graph} which in our test cases commonly took longest and was prohibitive beyond 20k data points. We tested implementations based on the python packages scikit-learn and scipy for this task and found the distance computation in scikit-learn to be faster. Rangesets were computed for five bins and averaged across variables. Fig.~\ref{fig:benchmark} illustrates the range of timings for the ten variables. Depending of the distribution of points and the discretization, we observe variations of 30\% of the mean value. Overall the plot shows that rangesets scale linear with the number of data points. Up to 5k data points we found a live computation of the rangesets acceptable (about 7sec waiting time). With more data points we recommend to downsample the dataset or precompute the rangesets.

Regarding the numbers of attributes, we were able to successfully use NoLiES for up to 16 attributes. Fig.~\ref{fig:betterlife} shows the GUI for a dataset with 11 attributes. To help users comprehend the connection between many attributes interactive selections are shared between all plots. To not obscure the underlying information, the outline curve is offset by edge width~\cite{FAROUKI199083}. In Fig.~\ref{fig:betterlife} the gray outline was manually drawn, optimized with the rangeset algorithm, and shared between all plots. In this way, the user can now go through all attributes, directly find the region of interest, and deselect irrelevant plots to focus on fewer attributes.

To help the user assess truthfulness of the projection we automatically include projection quality~\cite{motta2015graph} as auxiliary attribute in the GUI. 

%-------------------------------------------------------------------------

\begin{figure}
    \centering
    \includegraphics[height=4cm]{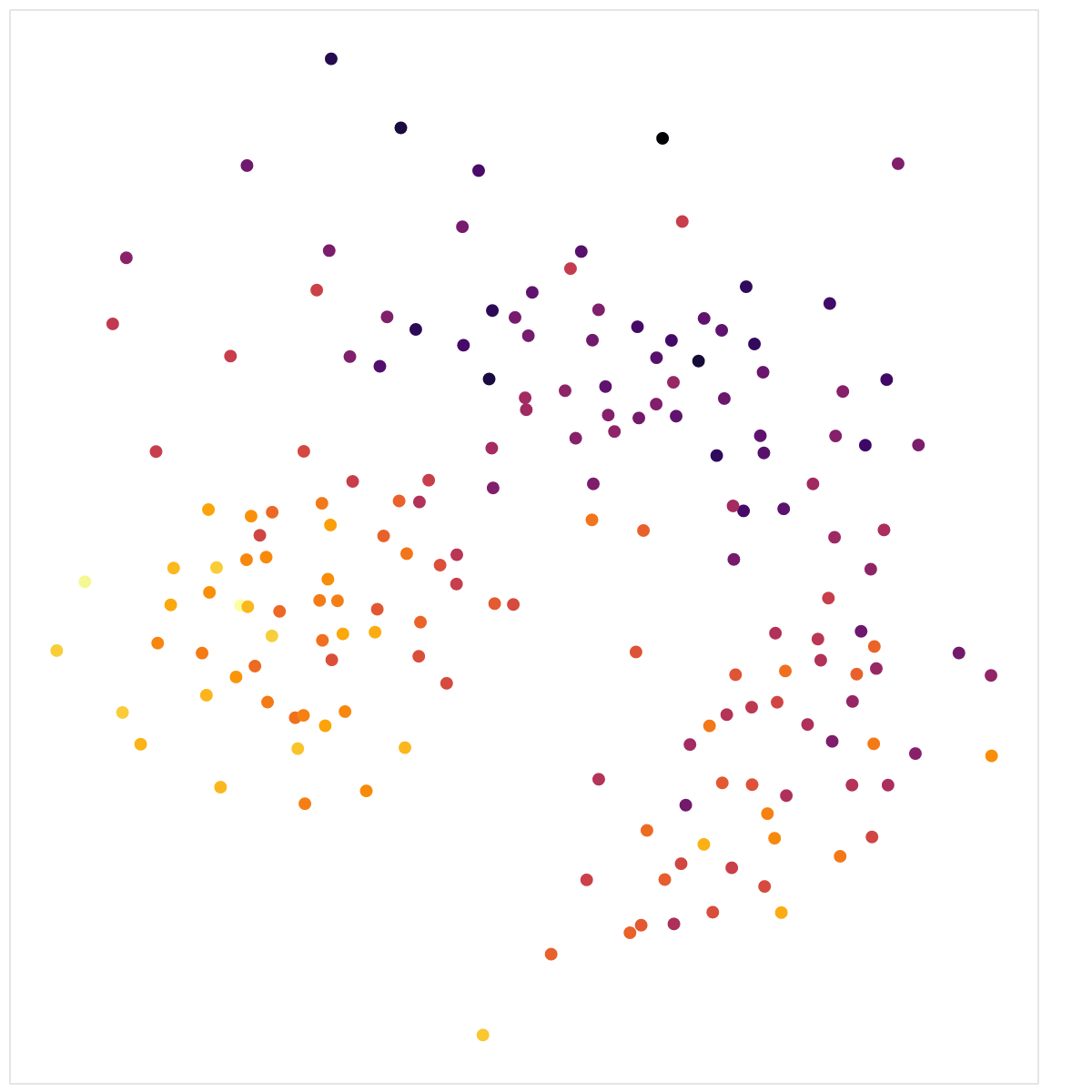}
    \includegraphics[height=4cm]{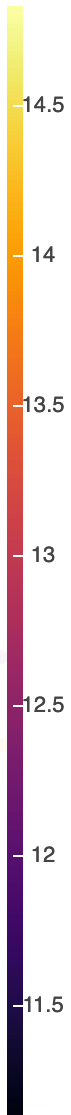}
    \includegraphics[height=4cm]{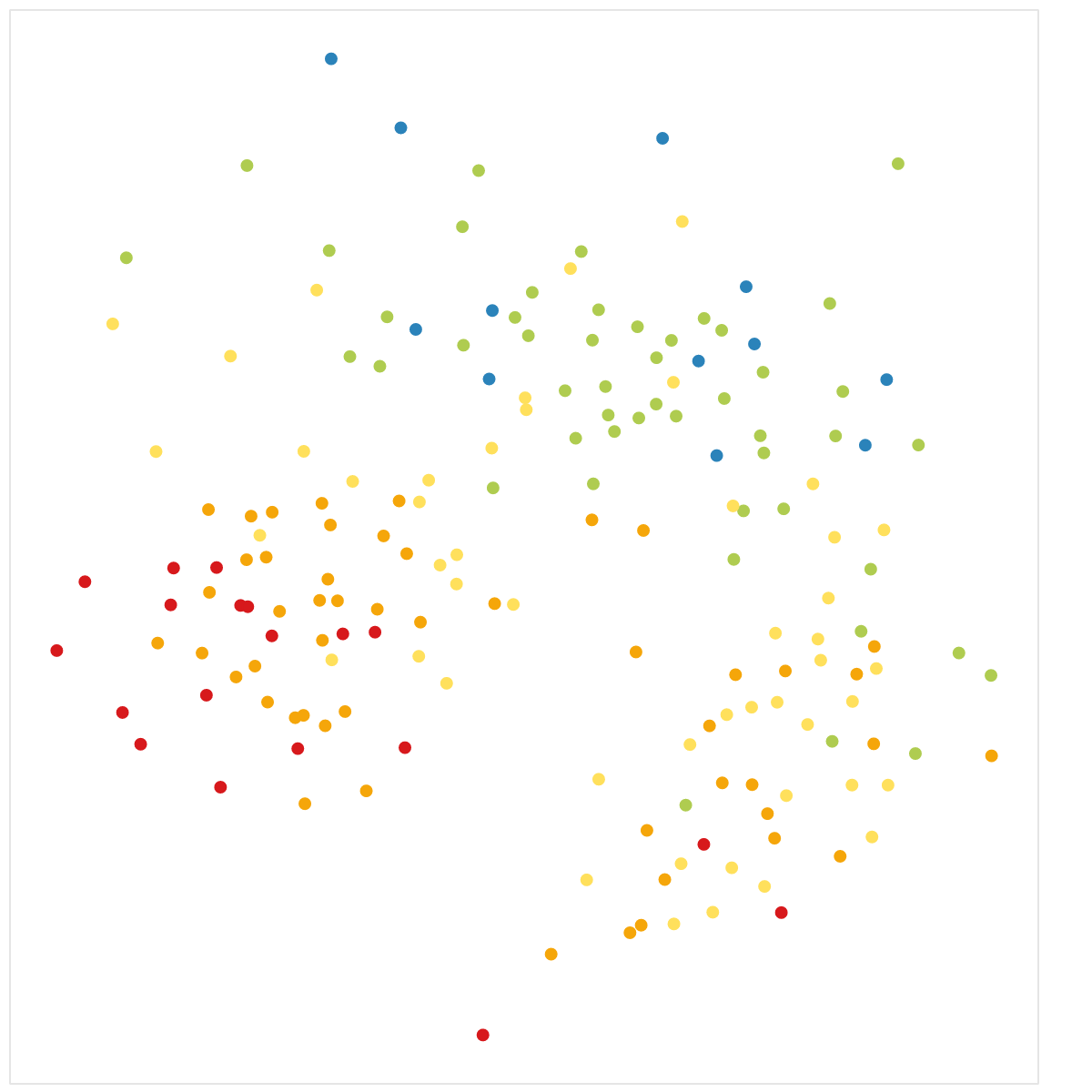}
    \includegraphics[height=4cm]{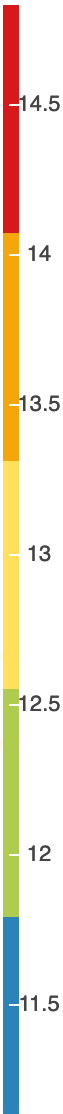}
    
    (a) Comparison of continuous (left) and discrete (right) colormap.
    \medskip
    
    \includegraphics[height=3.5cm]{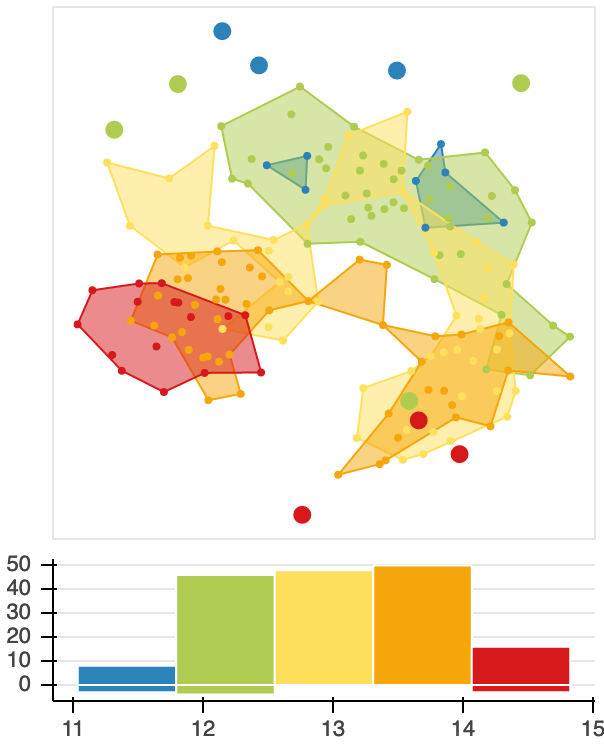}
    \includegraphics[height=3.5cm]{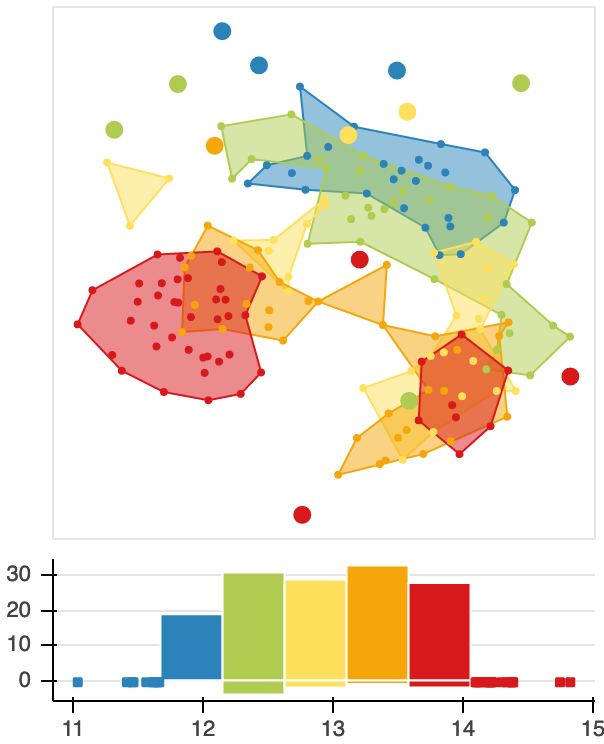}
    \includegraphics[height=3.5cm]{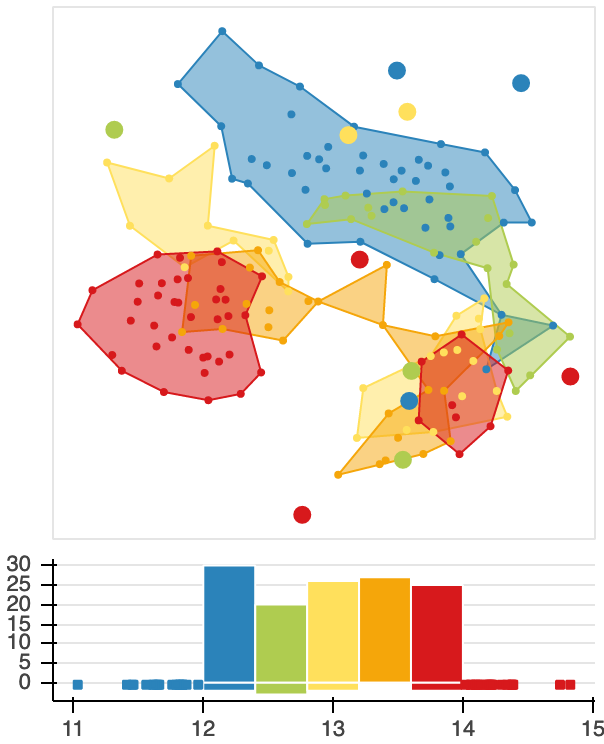}
    (b) Effects of changing range boundaries.
    \medskip
    
    \caption{Effects of discretization: (a) Color-coded glyphs using Matplotlib's perceptually uniform plasma colormap vs a discretized colormap based on Bokeh Spectral5. (b) Using the discretized colormap from (a) with additional contour outlines based on rangesets. The range can be interactively adjusted and we present three examples: (min,max), (11.7, 14.1), (12,14). In our example, we always include values below the minimum and above maximum in the extremal bins.}
    \label{fig:discretization}
\end{figure}

\subsection{Attribute Range Discretization}\label{sec:method:discretization}
\begin{figure*}
%    \begin{minipage}{0.4\textwidth}
%  \includegraphics[width=\linewidth]{figures/betterlife_1e.png}
%  \end{minipage}
%  \hfill
%  \begin{minipage}{0.2\textwidth}
%  \includegraphics[width=\linewidth]{figures/betterlife_2f.png}
%  
%  \includegraphics[width=\linewidth]{figures/betterlife_lifesat_histo.png}
%  \end{minipage}
%  \hfill
%  \begin{minipage}{0.35\textwidth}
%    \includegraphics[height=2.7cm]{figures/betterlife_3f.png}
%    
%    \includegraphics[height=2.7cm]{figures/betterlife_4f.png}
%  \end{minipage}
\includegraphics[width=\linewidth]{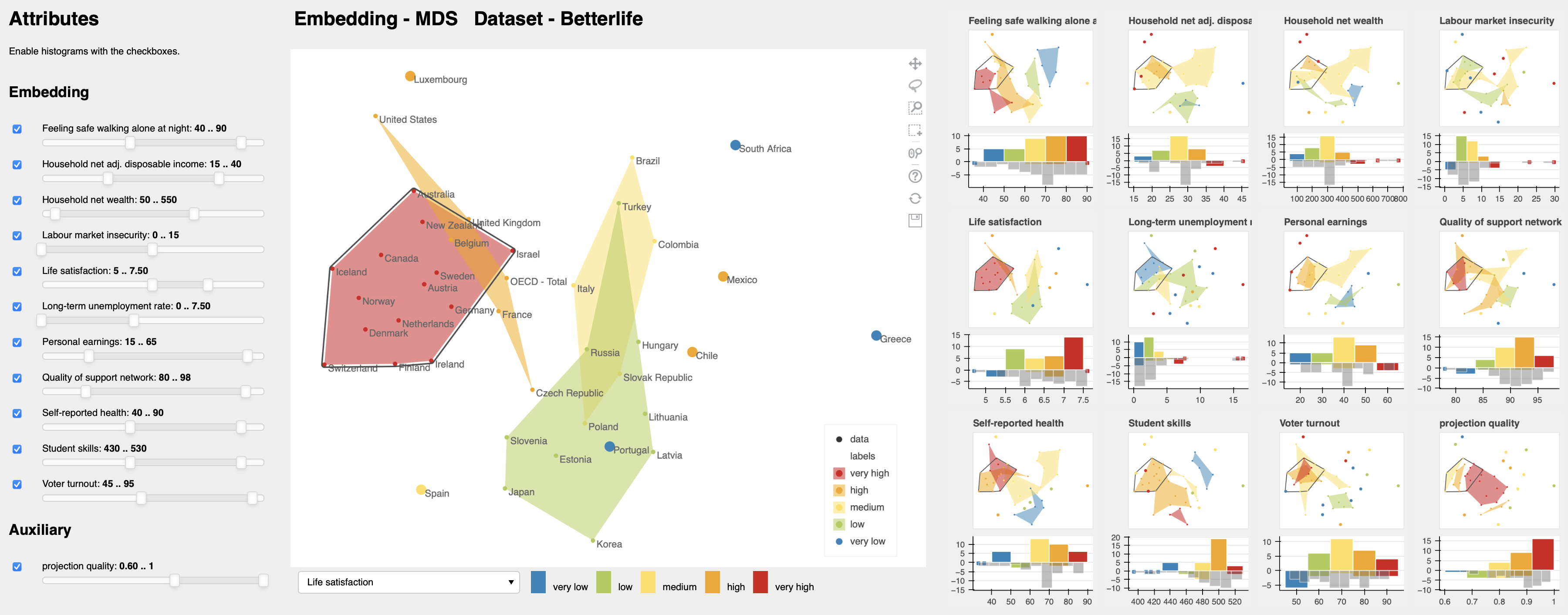}
\caption{How happy are you? The MDS embedding of the OECD Better Life data reveals multiple clusters that partially align with self perceived life satisfaction. The small multiples reveal patterns in six of the original attributes. Countries in the cluster of very happy countries (gray outline) have in common that they have no category (attribute) with very low ratings. Countries with very low life satisfaction (Greece, South Africa, Portugal) feature each their own challenges (i.a. unemployment, health, and/or safety).}
\label{fig:betterlife}
\end{figure*}

The contour algorithm described so far computes the contour for a single bin of the attribute histogram and, hence, data binning plays an important role for the rangesets. Histogram computation has been intensively studied in the statistics community and we can draw from excellent prior research~\cite{stats}. We use the numpy histogram function with settings for the following parameters: min-max-range of the attribute that is considered, number of bins for the discretization, step size of the discretization (uniform vs. non-uniform), and handling of values outside the selected range. We provide default values for all attributes following the reasoning below that can be overwritten by the user in the Jupyter Notebook. A screenshot of the entire system including small-multiples with histograms is given in Fig.~\ref{fig:betterlife}.
%As we want to discern attribute range location in the plane using these contours, we need to discretize the attribute value range. This concept is similar to the discretization and stacking approach as applied by horizongraphs~\cite{heer2009sizing} and supported by the findings of Kraus et al.~\cite{Kraus2020Assessing2A} who showed that reading continuous scalar field is a difficult task. 
%To control the discretization and overlaying, the following parameters need to be provided: min-max-range of the attribute that is considered, number of bins for the discretization, step size of the discretization (uniform vs. non-uniform), handling of values outside the selected range, color-coding of the bins/contours.

Initially, min-max-ranges for each attribute are defined by the extremal values in the data. Adjustments may be necessary to obtain bins with easy to read values or to account for the range of possible values (e.g. 0-100\% when values only range from 37-82\%). In NoLiES, this can be done interactively via the range-sliders below each attribute in the attribute panel (see Fig.~\ref{fig:betterlife}~(left) for an example). Upon dragging the slider, the contours update automatically giving instant feedback on the effects of the changes. Three samples of this interaction process on the wine dataset (attribute alcohol) are depicted in Fig.~\ref{fig:discretization}b. Like in this example, we experienced in general that the contours were fairly stable and the exact choice of the absolute range was not too critical. Customized ranges can be stored in the notebook and are automatically used in the next iteration.
%In our  implementation, the min-max-ranges per attribute can be  controlled by the user and can be interactively adjusted in the GUI (sec.~\ref{sec:system:gui}). By default they are set to the min-max-values of the attribute. 
Automatic outlier detection~\cite{wilkinson2017visualizing} and choosing thresholds with human-interpretable values~\cite{talbot2010extension} may be used to further automize the process.
%Binning the attribute range results in k parameters that need to be specified. The (min,max)-range of the histogram boundary and  the number of bins.

We discretize the selected range in the default case into five bins with equidistant boundaries. In our examples we found five bins sufficient to model any underlying distribution. Fig.~\ref{fig:betterlife} depicts the histograms for five bins in color and the version as computed by \texttt{numpy.histogram} below in gray. Using more contours resulted in hard to read images which aligns with the findings by  Kraus et al.~\cite{Kraus2020Assessing2A}, who report for continuous scalar fields  difficulties in precise readings of continuous maps on scatterplots, which aligns with findings by Tory et al.~\cite{tory2007spatialization}. A comparison of a perceptually uniform continuous colormap and our discrete one are given in Fig.~\ref{fig:discretization}a. We are aware that "rainbow"-colormaps are not a popular choice. However, the colors in this colormap are easy to name and highly distinct, which makes comparison in a small-multiples setting (like ours) easy and allows for easy and unambiguous communication. Hence, we assign each of the five bins a fixed color [blue, green, yellow, orange, red] and a label [very low, low, medium, high, very high]. Again the colormap can be easily changed to personal preferences in the notebook and the Bokeh-library offers a rich set of default colormaps to choose from. We explored non-uniform discretization to better adapt to local structures in the point cloud, but found this to be misleading when interpreting the color distribution. %We use the spectral colormap to reflect the notion of the Likert-scale.
% All parameters can be easily adjusted in the jupyter notebook to personal preferences and application requirements as will be demonstrated in the use case on thermodynamics where 20 classes had to be incorporated into the design.

An additional augmentation of the histograms in NoLiES is the use of the negative y-axis. Here, we encode data points that are outside the ranges. Datapoints outside the min-max-range are encoded as glyphs below the y-axis. Datapoints that are outliers for a particular rangeset (i.e., their scalar value falls within one of the bins, but their distance to the nearest point is larger than $\epsilon$) are counted and encoded as bars extending in the negative y-direction. In this way, the user can directly see how parameter settings affect the topology of the rangeset chart.

%\textbf{continuous vs discrete}: 
%design criteria
%- few parameters that are easy to control
%- ease of implementation
%- works for few and massive points
%- easy to explain
%- little visual overload - plain and simple charts

%reasons for discretization
%- non-linear projections may lead to hard to comprehend value distribution with many subtle bends and tilts. Discretization simplifies the structure and makes data easier to comprehend.
%- We chose five levels similar to horizon graphs. This goes with human intuition. reference?
%- Topology may help to find optimal discretization. For the time being we go for human interaction.

%-------------------------------------------------------------------------

\subsection{Visual Encoding}\label{sec:method:encoding}
To visually represent the rangesets, we augment the scatterplot with polygons and color-coded glyphs. The contour for each bin is rendered as a filled transparent polygon (alpha value is 0.5) and is rendered below the scatterplot. We additionally render (a subset of) the datapoints including those inside contours, to provide the user with a sense of data density. 
%The visual grouping additionally increases the signal-to-noise ratio as the chart provides a high-level abstraction of visual patterns~\cite{chen2010information}. 
Custom contours (gray outline) can be added to highlight a selected group of points. To highlight outliers, we chose the channel glyph size increasing the radius of the point glyphs. Additionally, outliers are plotted on top of the contours to not be hidden by multiple transparent layers. In the histogram view, we highlight outliers by counting them on the negative side of the y-axis.

%- Outliers become more easily visible; we also increase signal-to-noise ratio by this clear distinction, points are either inside or out and humans do not have to decide for each point.

%We considerd the following two filtration parameters as are common in computational geometry: triangle area and maximal edge length of the triangle. problem with the area is that we keep long spiky triangles. Hence, we go for edge length
%\subsection{Attribute Traces}

%-------------------------------------------------------------------------
%-------------------------------------------------------------------------

\section{Interactive System}\label{sec:system}
NoLiES is implemented in python using the panel library for interaction support (Fig.~\ref{fig:workflow}). This allows to run the application both in the browser as a Jupyter notebook and as a standalone application. In the analysis process, we alter between the two views using each where it is strongest. The software is written in modules. Generic code like the rangeset computation is stored in an extra module. We create one notebook per dataset and can use it in this way as a storage location for knowledge that was derived during the analysis process.
We intentionally do not include too many settings in the GUI to make it easy to use and provide places to store information in a uniform way.

%-------------------------------------------------------------------------

% \subsection{Support in the Jupyter Notebook}\label{sec:system:notebook}
\textbf{Jupyter Notebook}: The notebook provides three sections that guide the user through the analysis process. Part 1 handles data loading, preprocessing, and cleaning. Additionally, we include placeholders for commonly used system parameters like custom ranges for sliders, attribute filters, and the rangeset threshold. Part 2 handles the multidimensional projection. Multiple widely used dimensionality reduction methods as implemented in the scikit-learn library~\cite{scikit-learn} are included and can be selected by the user. We also include best practices for dimensionality reduction like attribute scaling and tests for correlation. Part 3 takes care of the visual design and interactions. The system is implemented in python and can be rapidly manipulated by users with limited programming experience as we found in an informal user study (see Discussion).

%-------------------------------------------------------------------------

%\subsection{Interactive NoLiES}\label{sec:system:gui}
\textbf{Interactive App}: The GUI of NoLiES consists of three major components (Figure~\ref{fig:workflow}): (i) An attribute view that lists all included attributes from the raw data, their ranges, and the selected sub-ranges to be used for the binning. (ii) The embedding renders the projected data as a point cloud with optional labels. The user can interactively alter the displayed rangesets in a dropdown menu. The title of the chart automatically includes the applied projection method as defined in the notebook. (iii) The small multiples view provides a quick overview over all selected attributes distributions and present the histograms for the binned attributes. Views can be interactively switched on and off in the attribute view with a checkbox. 
The attribute sliders are interactive and upon moving the sliders, the outlines and the histograms are updated interactively. This procedure helps to understand attribute value locations and the effect of the discretization. The user can also use this technique to manually filter for outliers and set tighter value bounds for the displayed contours. Once they found good default values, these can be stored in the notebook and will be set at defaults in the next run.

%Another important feature that we integrated is connecting all data views with respect to selections. Often this is done by color or alpha-value manipulations, but these channels are already heavily used in our design and are no longer salient. Hence, we integrated an additional outline curve for selected data points (gray outline in Figure~\ref{fig:betterlife}). To not obscure the underlying information, the outline curve is offset by edge width~\cite{FAROUKI199083}. Using this outline, the user can quickly compare selections across multiple views.

%-------------------------------------------------------------------------

% \subsection{Implementation}
\textbf{Implementation}: NoLIES is implemented in python in the Jupyter Notebook. The GUI is realized using the panel library~\cite{panel} and charts are created using bokeh~\cite{bokeh}. Geometric operations are realized with the shapely library~\cite{shapely}. Multidimensional projections and data preprocessing are provided through scikit-learn~\cite{scikit-learn}. Data handling is realized using pandas~\cite{scipy}. Topological data analysis is implemented in NoLiES and supported through methods in the shapely library~\cite{shapely}. NoLiES is available on GitHub\footnote{https://github.com/leitte/NoLiES}.

%-------------------------------------------------------------------------
%-------------------------------------------------------------------------

\section{Case Studies}\label{sec:casestudies}
In the following, we present three case studies with increasing complexity. The Better Life dataset is easy to comprehend and follow. %The wine dataset is widely studied and comes with complex patterns and a moderate number of data points. 
The forest covertype dataset contains more than 4000 data points and is difficult to project. The real-world study in thermodynamics targets explainable machine learning and large number of contours in the rangeset.

\begin{figure*}
    \centering
    \begin{subfigure}{.2\linewidth}
        \includegraphics[height=3cm]{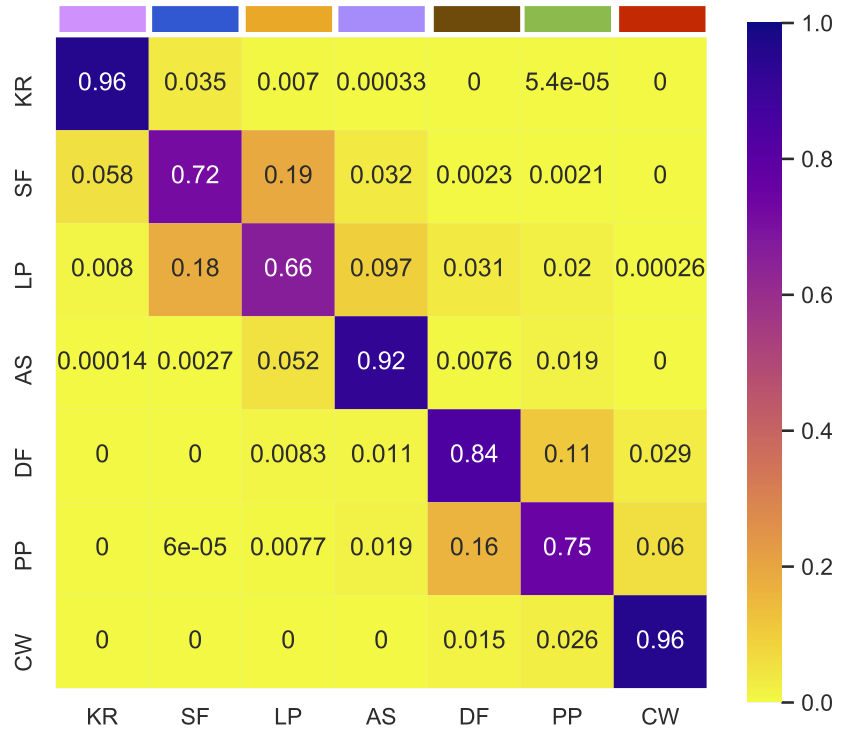}
        \caption{Confusion matrix}\label{fig:covtype:confusion}
    \end{subfigure}
    \begin{subfigure}{.34\linewidth}
        \includegraphics[width=\linewidth]{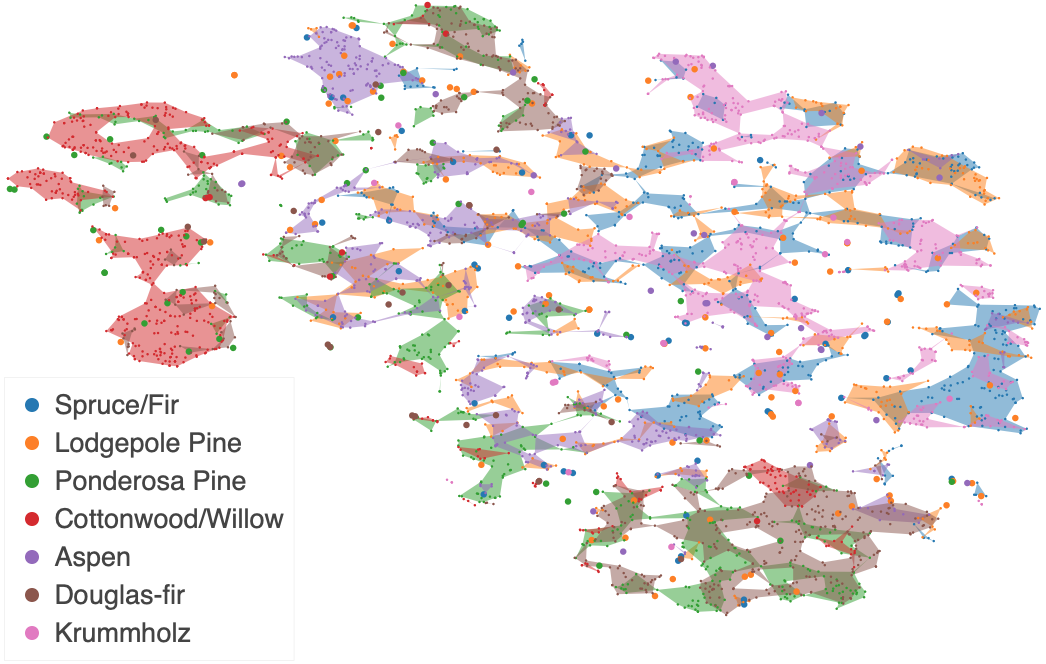}
        \caption{Forest type}\label{fig:covtype:type}
    \end{subfigure}
    \begin{subfigure}{.35\linewidth}
        \includegraphics[width=\linewidth]{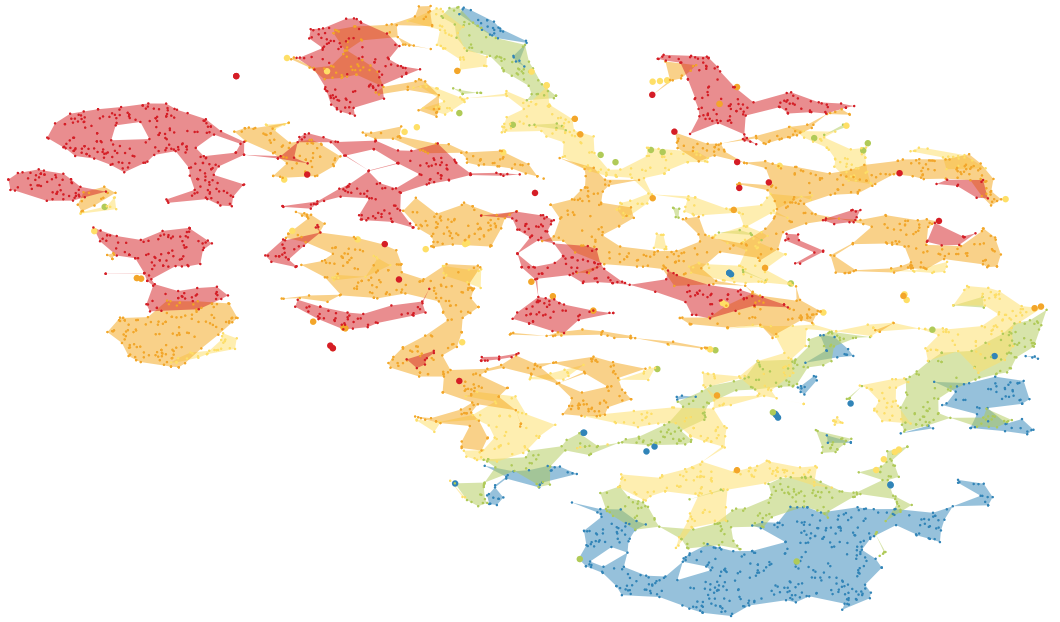}
        \caption{Shade at 9am}\label{fig:covtype:elevation}
    \end{subfigure}
    %
    % \begin{subfigure}{.225\linewidth}
    %     \includegraphics[width=\linewidth]{figures/covtype600_elevation.png}
    %     \caption{Elevation}
    % \end{subfigure}
    % %
    % \begin{subfigure}{.225\linewidth}
    %     \includegraphics[width=\linewidth]{figures/covtype600_shade9.png}
    %     \caption{Shade at 9am}
    % \end{subfigure}
    \caption{Forest covertype dataset with 4.200 data points and 9 attributes: 70\% classification accuracy can be achieved for this dataset~\cite{Blackard1999ComparativeAO} with common confusion of spruce/fir and lodgepole pine (blue vs orange) as well as douglas-fir and ponderosa pine (brown vs green) (a) which is illustrated by overlapping contours in the embedding (b). (c) Original attributes help characterize classes and analyze the algorithm.\vspace{-2mm}}
    \label{fig:covtype}
\end{figure*}    
\subsection{OECD Better Life}
The OECD Better Life dataset~\cite{oecd} measures 25 attributes for 40 countries (+ OECD mean). The goal is to understand which factors promote a society's well-being. For illustration purposes we chose 11 attributes that covered the general mix of topics, had varying attribute distribution profiles, and did not alter the general structure of the projected data too much.
Fig.~\ref{fig:betterlife} shows the entire GUI view with all attributes enabled in the small multiples view. 
%The MDS projection is annotated with country labels and we can observe two prominent clusters in the left and bottom part of the chart. USA and Luxembourg form an outlying group, as do the countries in the upper-right corner. Greece is drawn at a large distance from all other countries, which indicates its high dissimilarity to all countries.

The central chart renders the rangeset for self-reported life satisfaction (range (0,10)). The histogram on the left tells us that values range from 4.7 (South Africa - requires interaction to be determined) to 7.6 (Denmark, Finland, Norway) which were discretized into bins of width 0.5 ranging from satisfaction values of 5 to 7.5 (out of 10). Countries outside the histogram range are included in the two extremal bins. The small chart on projection quality shows that the MDS projection works in general very well and can retain the neighborhood structure (except for Luxembourg). We can observe that life satisfaction strongly correlates with the visual clusters in the 2D embedding and that the red and green class (very high and low happiness) contain most data points. The very unhappy countries (blue points) are spread across the plot. They are not grouped in a rangeset contour, but form all outliers which is reflected in the histogram (bar below the zero-line). We also observe some outliers in the orange class (large dots in the scatterplot; Chile and Mexico) who report high happiness, but are in many categories close to more unhappy countries.

%interactive user interface for the OECD Better Life dataset~\cite{} which we will analyze in the following. The overview depicts the distribution of the embedded data which was projected using multi-dimensional scaling (MDS). There is one prominent cluster visible (green in Fig.~\ref{fig:betterlife_overview}), two less pronounced clusters and some points with large distances to other points (included in clusters in Fig.~\ref{fig:betterlife_overview}). Figure~\ref{fig:betterlife_overview} depicts the result of an automatic clustering and the goal is to use the embedding to better explain this dataset. 

%\begin{figure}[t]
%    \centering
%    \includegraphics[width=.9\linewidth]{figures/wine_b1.png}
%    
%    \includegraphics[width=.9\linewidth]{figures/wine_b2.png}
%    
%    \includegraphics[width=.9\linewidth]{figures/wine_b3.png}
%    \caption{MDS-embedding of the wine dataset: (top) The ground truth contains three classes representing cultivars (red, yellow, blue). %The red class is selected and the outline synchronizes in all plots. Red-class wines have high values in alcohol content, flavonoids, hue, and proline.
%    \vspace{-4mm}}
%    \label{fig:wine}
%\end{figure}

%To inspect correlations with other attributes
To obtain an overview over the distribution of assessed attributes, we first look at the small multiples display in Figure~\ref{fig:betterlife}~(right).
%individual attribute value distributions. 
 Attributes with a blue checkbox are presented in the small multiples summary on the right (here all of them). We observe that the attribute `Feeling safe walking alone' decreases from bottom-left to top-right. The histogram below the small multiple depicts the attribute distribution. We see that countries colored in red have $80-90\%$ agreement and blue countries less than $50\%$, i.e., people in countries contained in the red polygons agree to $80-90\%$ with the statement that they feel safe when walking alone. Other attributes like `Labour market insecurity' and `Student skills' feature a more complex and harder to describe distribution of attribute values, which is expected in an MDS embedding. %In section~\ref{} we will compare different augmentation strategies to communicate attribute value distributions and report feedback from the expert user study.

\begin{figure}[t]
    \centering
    \begin{subfigure}{.47\linewidth}
        \includegraphics[width=\linewidth]{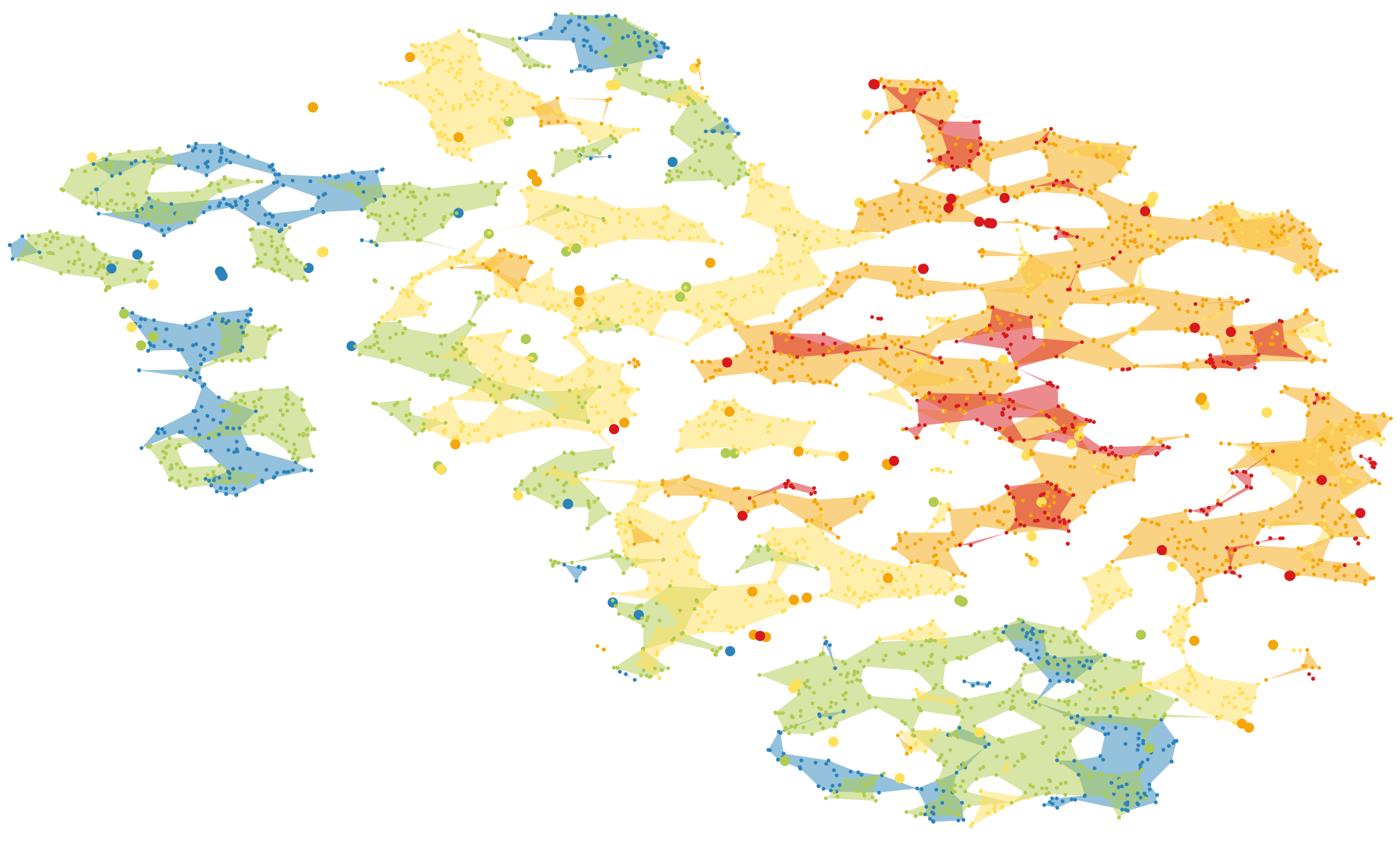}
        \caption{Rangesets}
    \end{subfigure}
    \begin{subfigure}{.47\linewidth}
        \includegraphics[width=\linewidth]{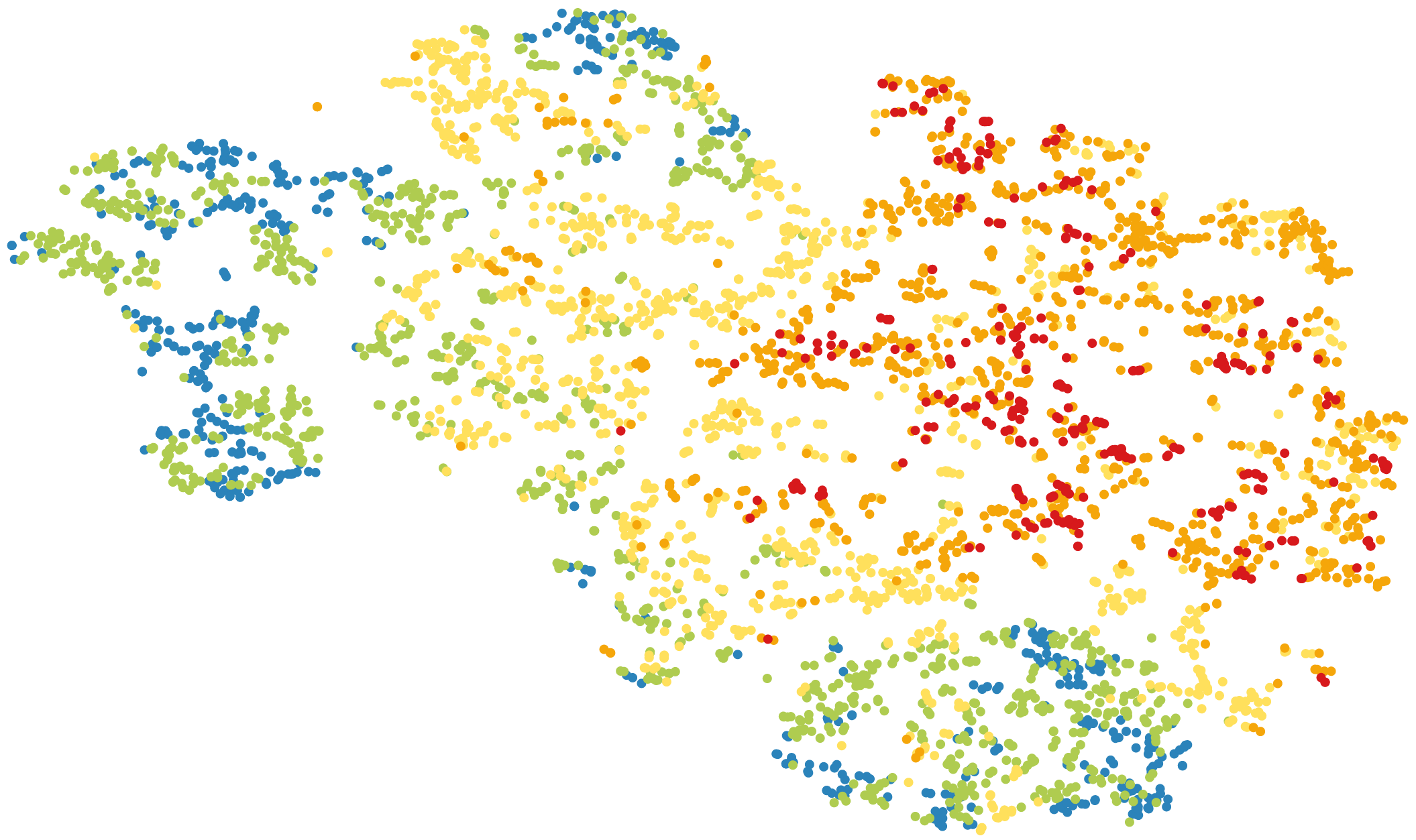}
        \caption{Color-coding}
    \end{subfigure}
    \caption{Comparison of augmentation techniques: (a) Rangesets quickly outline regions of different parameter values. The size of outliers can be interactively increased. (b) Augmentation based on glyph-based color coding.\vspace{-4mm}}
    \label{fig:covtype_compare}
\end{figure}

Next we focus on the region outlined in gray, which contains all countries with very high life satisfaction. The outline was drawn manually using the selection tool. %Selections are directly supported in Bokeh and can be forwarded to other plots. 
%There are some countries with happy people outside this cluster (Mexico, Chile, Czech Republic) which we will analyze later. Selecting the green countries in the overview chart creates a black outline that is forwarded to the small multiples. 
Concentrating on the colors inside the outline for each attribute, we can quickly observe that countries in the selection have diverse, but generally positive values in all attributes, except for labour market insecurity where low values are good. In summary, we can state that countries with very high life satisfaction do well in all assessed categories. This also discriminates the two major clusters (very happy (red) and unhappy (green) countries). Countries in the green cluster have at least one problematic category with low and very low values, disposable income, self-reported health, and safety being the most prominent ones.

The US and Luxembourg form their own small cluster close to the very happy countries, but only rate themselves with high life satisfaction. Comparing these two countries to colors inside the selection polygon for the very happy countries, we can identify attributes in which they diverge. Looking for color differences, we identify students skills (lower than in selection polygon) and household net income (higher than in selection polygon). From our investigation we can conclude that money alone does not make happy and that it probably is a combination of factors that result in (only) high life satisfaction in these two countries.

\subsection{Forest Covertype}
The forest covertype dataset~\cite{Blackard1999ComparativeAO} covers 581k datapoints and 54 cartographic attributes like elevation, slope, and shade. The goal is to predict the forest type (7 classes) using these attributes only. Blackard et al.~\cite{Blackard1999ComparativeAO} report 70\% accuracy using an ANN. The misclassification as depicted in the confusion matrix (Fig.~\ref{fig:covtype:confusion}) are extracted from their paper. Rows in the confusion matrix correspond to actual classes and columns to the predicted ones. We observe major misclassification between spruce/fir (SF) and lodgepole pine (LP) as well as douglas-fir (DF) and ponderosa pine (PP). Multiple other misclassifications occur (mainly close to the diagonal) and some classes are never confused, e.g. CW with KR/SF/LP/AS.

For the dimensionality reduction we sampled 600 data points from each of the 7 classes resulting in 4,200 data points. The embedding was computed using the 10 numeric attributes. The discarded 44 categorical attributes distinguish between wilderness areas (4 attributes) and soil types (40 attributes). The t-SNE-based embedding (perplexity: 30, early exaggeration: 30) is depicted in Figure~\ref{fig:covtype:type}. 
%For large numbers of datapoints the running time of MDS becomes too long ($\mathcal{O}(n^2)$ running time for the default implementation) and we also wanted to demonstrate an alternative projection technique. 
The rangeset coloring encodes the ground truth class labels. We observe that there is a lot of overlap between colors. On closer inspection, we find that only certain colors overlap, which is in agreement with the confusion analysis of the ANN model. We observe overlap, e.g., between blue/orange/pink for SF/LP/KR or brown/green for DF/PP. Additionally, we observe that some colors only occur in particular regions, e.g. red areas are located at top-left and in multiple small regions on the southern boundary of the purple region. 

It is important to note that structural analysis of t-SNE plots is challenging and may easily lead to misreadings~\cite{wattenberg2016how}. The overlay with rangesets can help to counteract common misperceptions. As stated by Wattenberg, cluster sizes mean nothing in t-SNE~\cite{wattenberg2016how}. With the overlayed attribute-based rangesets, the user can reconstruct the underlying distances between datapoints. Figures~\ref{fig:covtype}c+\ref{fig:covtype_compare} augment the plot with two of the original attributes. Regions in the same color are close in high-dimensional space with respect to this attribute. For example, low elevation regions are split into two groups by t-SNE (top-left and bottom-right). For the red Cottonwood/Willow covertype, we can thus deduce that they mainly grow in areas with low elevation and high shade values at 9am. These observations can also be made using classical glyph-based color coding. We found in our experiments that the trust in the observations was higher using rangesets -- a detailed elucidation, however, is subject to future studies.

%600 * 8 data points, 9 attributes, tSNE

%concept: covers many data points and attributes; challenging classification problem as in paper~\cite{Blackard1999ComparativeAO}; ANN achieves accuracy of 70.5\% with confusion as illustrated in Fig.~\label{fig:covtype:confusion}. Major confusion between SF/LP and DF/PP which can also be seen in the tSNE embedding with respective coloring Fig.~\ref{fig:covtype:type}: we have substantial overlap between these colors. Additionally we see that there is little to no overlap for forest types that are not confused eg red and pink. We can also characterize the regions where trees are growing, here we see elevation and shade. CW and DF grow in higher regions (high elevation) and LP-pink in low regions. Shade has a different profile. The color also helps to understand tSNE which has a highly unpredictable transformation.

%overview figure~\ref{fig:covtype}

\begin{figure}
    \centering
    \includegraphics[width=\linewidth]{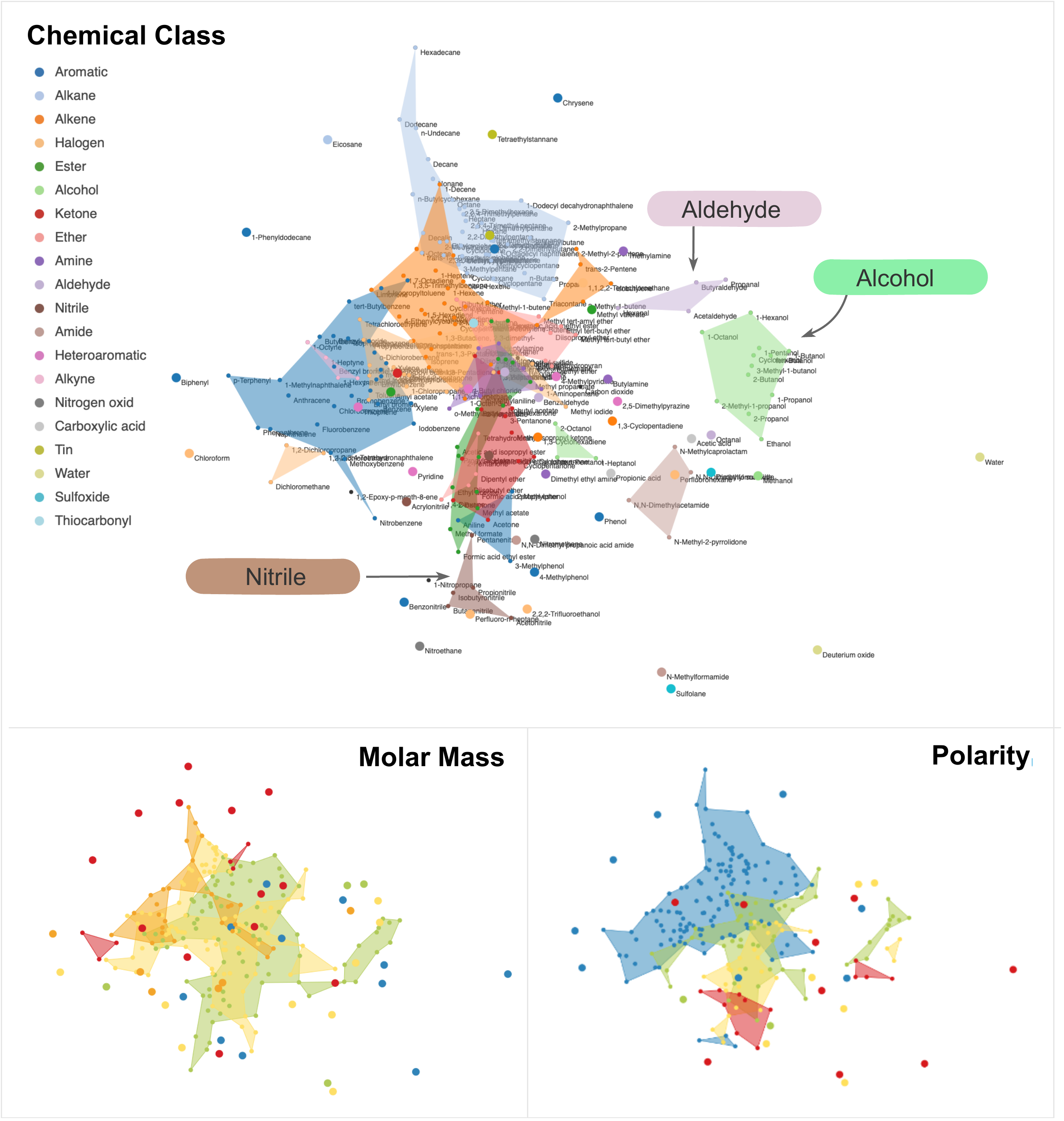}
    \caption{MDS-embedding of the four MCM features of 240 solutes trained to data for activity coefficients in binary mixtures~\cite{doi:10.1021/acs.jpclett.9b03657}. The ground truth as captured by domain knowledge covers 20 chemical classes (color code, top-left) and information on the molar mass and polarity (bottom). Comparing domain knowledge and latent MCM features $u_i$ (Fig.~\ref{fig:teaser}) helps explain black-box machine learning techniques.}
    \label{fig:thermo1}
\end{figure}

\subsection{Matrix Completion in Thermodynamics}\label{sec:case:thermo}
The prediction of fluid properties plays a central role in chemical engineering, e.g., for process design and optimization, since experimental measurements are usually cumbersome and expensive. Methods to predict the properties of binary mixtures are of particular importance, since also the properties of multi-component mixtures can often be described based on information on the binary `submixtures'~\cite{carlson1942vapor}.  
Matrix completion is a novel promising ML approach for this purpose~\cite{doi:10.1021/acs.jpclett.9b03657,D0CC05258B}. However, while data-driven matrix completion methods (MCM) yield great performance in predicting fluid properties, they are not intuitive and therefore difficult to understand from a physical perspective. This strongly reduces confidence among engineers and natural scientists and hampers their application. Hence, tools that enable a physical interpretation of MCM are paramount.
In this case study, we apply NoLiES for this purpose. Note that alternative approaches using glyph coloring and clustering failed in communicating any relationship between latent MCM features and domain knowledge (Fig.~\ref{fig:teaser}). Figure~\ref{fig:thermo1} shows MDS projections of four MCM features of 240 solutes, which were trained to experimental data for the activity coefficients of these solutes at infinite dilution in 250 solvents (at a temperature of 298.15~K)~\cite{doi:10.1021/acs.jpclett.9b03657}. Hence, these MCM features constitute latent descriptors of the solutes that are derived from mixture properties (here: activity coefficients) only.  

The embedding features multiple clusters, which we overlay with expert knowledge on the chemical classes the solutes are allocated to (denoted by the color code of rangesets and glyphs) in Figure~\ref{fig:thermo1}(top) and observe a surprising coherence. We learn that the chemical class of a solute is a suitable descriptor regarding the solute's MCM features (and its activity coefficients in mixtures to which the MCM features were trained). We furthermore observe that some chemical classes are very characteristic, e.g., alcohol (right, light green), aldehyde (top right, light purple), and nitrile (bottom, dark brown), whereas the contours of others strongly overlap (mainly located in the center). Also note that water and deuterium oxide (heavy water) appear as exceptional solutes (bottom right edge, light olive green), which fits well with their exceptional macroscopic properties. In addition, correlations of the MCM features with other physical descriptors, such as the solute’s molar mass and polarity
(Figure~\ref{fig:thermo1}(bottom)), are found.

Looking at the rangesets for the four MCM features (Fig.~\ref{fig:teaser}(bottom)), we observe clear spatial structure. A detailed analysis of the link between MCM features and physical properties of the solutes is subject to future research and not within the scope of this paper. However, the results shown here indicate that NoLiES offers exciting physical insights into latent MCM features, which will serve as basis for a targeted enhancement of MCM, e.g., for selecting suitable physical descriptors to support the data-driven approach or predicting the MCM features of additional solutes based on readily available information on the solutes.

\section{Discussion and Future Work}\label{sec:discussion}
The goals that we set out to accomplish were as follows: Design a technique and system (i) that is easy to use and comprehend, (ii) that is applicable to all types of embedding techniques, (iii) that can be directly integrated into existing analysis pipelines, and most importantly (iv) that enables the user to quickly and correctly understand attribute value distributions in the embedded data. Goals (i + iv) are demonstrated in the use cases and were assessed in an informal expert user study with five domain scientists from various application fields that need to interpret high-dimensional data. We demonstrated NoLiES and gave them access to the notebooks. They all used data from their own work and explored structure and outliers therein (one application is reported in sect.~\ref{sec:case:thermo}). All experts commented directly that the visualization is visually appealing and easy to comprehend. One user commented that rangesets reminded him of cartography, which we deem an interesting analogy. The interactive GUI part proved directly accessible to all levels of computer literacy. The users with background in programming/python were additionally able to download NoLiES from GitHub and customize the notebook for the exploration of their own data. Goals (ii + iii) are ensured by the implementation in Python and Jupyter Notebooks and the lack of coupling to the embedding. We demonstrate rangesets for MDS and t-SNE, which are often used to reveal inherent structure and clusters in the data. We also tested rangesets on PCA and SVM projections with similar results. Rangesets are computed in a post-processing step and the code can be easily used outside of NoLiES.

Limitations that we encountered relate mainly to scalability issues. With increasing number of data points interactivity slows down. Computing contours in the forest covertype dataset with 4k data points takes about one second, which also holds for the computation of the embedding on a regular desktop PC. For both routines, we use external libraries that we cannot easily improve. As rangesets and the embedding are computed only once, we found the latency acceptable. Similarly, reading rangesets became challenging with many classes as in the thermodynamics example with 20 chemical classes. We currently use a bokeh default colormap. Colormaps optimized for perception that are aware of spatial overlap may further increase visibility~\cite{Mayorga2013SplatterplotsOO}.

The geometrical nature of rangesets directly suggests several extensions like the support of Boolean operations on the sets. This concept could also be applied for a further automation of the analysis process of cluster properties, which we currently did fully manually.

%-------------------------------------------------------------------------
\section{Conclusion}
In this paper, we presented NoLiES, an interactive system for the interpretation of embeddings of multi-dimensional data projections. We introduced rangesets, an augmentation strategy for embeddings that outline datapoints with similar values in multiple non-convex contours. Rangesets have a dedicated handling of outliers and their only parameter is the maximal acceptable edge length between connected points. We discussed the relationship between rangesets and algebraic topology and demonstrated how the theory can be used to control the rangeset parameter. To work with rangesets of multiple data attributes, NoLiES integrates an interactive small multiples concept that is linked by selections and color coding. Important knowledge obtained during the analysis of a dataset can be stored in the notebook and used in future analysis.

%We have implemented a prototype for exploratory analysis of high-dimensional datasets embedded in a 2D canvas via nonlinear DR. We achieve this by transforming the samples together with their attributes into an embedding in two-dimensional space which provides context for our analyses. Our analyses itself are not only aided by this context but also by additional information like the value distribution of attributes or the density distribution within intervals. Interactively one can define and re-define ranges of attributes of interest to reveal information, that would otherwise be hidden in large amounts data.

%We provide several fast and simple interaction possibilities for the user to interact with the visualization. This interaction helps to get a first understanding of underlying structures and the distribution of samples. One can identify clusters and how the attribute values are distributed over the embedding.

\section{Acknowledgements}
%Omitted for double blind review.
This research was funded by the Deutsche Forschungsgemeinschaft (DFG, German Research Foundation) – 252408385 – IRTG 2057.

\bibliographystyle{abbrv-doi}

\bibliography{bib-refs}
\end{document}